\documentclass[sn-mathphys-num]{sn-jnl}

\usepackage{graphicx}%
\usepackage{multirow}%
\usepackage{amsmath,amssymb,amsfonts}%
\usepackage{amsthm}%
\usepackage{mathrsfs}%
\usepackage[title]{appendix}%
\usepackage{xcolor}%
\usepackage{textcomp}%
\usepackage{manyfoot}%
\usepackage{booktabs}%
\usepackage{algorithm}%
\usepackage{algorithmicx}%

\usepackage{algpseudocode}%
\usepackage{listings}%
\usepackage{caption}
\usepackage{subcaption}

\usepackage{soul}

\usepackage{bbm}

\theoremstyle{thmstyleone}%
\theoremstyle{thmstyletwo}%

\theoremstyle{thmstylethree}%

\begin{document}

\title[Article Title]{Modeling Sampling Distributions of Test Statistics with Autograd}


\author{\fnm{Ali} \sur{Al Kadhim}}\email{aa18dg@fsu.edu}

\author{\fnm{Harrison} \sur{B. Prosper}}\email{hprosper@fsu.edu}

\affil{\orgdiv{Department of Physics}, \orgname{Florida State University}, \orgaddress{\street{600 W College Ave}, \city{Tallahassee}, \postcode{32306}, \state{Florida}, \country{USA}}}


\abstract{
Simulation-based inference methods that feature correct conditional coverage of confidence sets based on observations that have been compressed to a scalar test statistic require accurate modeling of either the p-value function or the cumulative distribution function (cdf) of the test statistic. If the model of the cdf, which is typically a deep neural network, is a function of the test statistic then the derivative of the neural network with respect to the test statistic furnishes an approximation of the sampling distribution of the  test statistic. We explore whether this approach to modeling conditional 1-dimensional sampling distributions is a viable alternative to the probability density-ratio method, also known as the likelihood-ratio trick. Relatively simple, yet effective, neural network models are used whose predictive uncertainty is quantified through a variety of methods.
}

\maketitle


\section{Introduction} 
Automatic differentiation (see, for example, Ref.\cite{autodiff}) has revolutionized machine learning, permitting the routine application of gradient descent algorithms to fit to data models of essentially unlimited complexity. The same technology can be used to take the derivative of these models with respect to their inputs without the need to explicitly calculate the derivatives, \cite{CHEN2020123344} which can be cumbersome or intractable for complex models. In this paper we leverage this capability to investigate whether it is possible to obtain an accurate
approximation of the 
probability density function (pdf),  $f(x \mid \boldsymbol{\theta})$, given a neural network model of the associated conditional cumulative distribution function (cdf), $F(x\mid\boldsymbol{\theta})$, using the fact that
\begin{align}
    f(x \mid \boldsymbol{\theta}) & = \frac{\partial F(x \mid \boldsymbol{\theta})}{\partial x} ,
    \label{eq:pdf}
\end{align}
where $\boldsymbol{\theta}$ are the parameters of the data-generation mechanism, which we distinguish from the parameters $\mathbf{w}$ of the neural network (NN) model.
A pdf $f(x \mid \boldsymbol{\theta})$  is the key ingredient in both frequentist and Bayesian inference, while the cdf arises naturally in the context of simulation-based frequentist inference \cite{cranmer2016approximating,Cranmer:2019eaq,Brehmer:2019xox,dalmasso2023likelihoodfree,ALFFI}.
Simulation-based inference (SBI) makes it possible to perform inferences without the need for explicit knowledge of the statistical model. However, given that SBI often furnishes the cdf $F(x \mid \boldsymbol{\theta})$, standard inferential methods may be brought to bare if one had access to the associated pdf. For example, it would be possible to construct confidence intervals for any parameter of the model or deploy methods such as maximum likelihood.

Equation\,(\ref{eq:pdf}) furnishes an approximation of the pdf $f(x \mid \boldsymbol{\theta})$ whether $x$ is a function of the underlying observations $D$ only or if $x = \lambda(D; \boldsymbol{\theta})$ is a test statistic that depends on $D$ as well as on the parameters $\boldsymbol{\theta}$. Moreover, computing the derivative of the cdf using automatic differentiation (or autograd) to obtain the pdf is exact; autograd does not use finite difference approximations. Our main contributions are as follows.
\begin{itemize}
    \item We suggest a novel use of automatic differentiation.
    \item We investigate the feasibility of accurately approximating the pdf by leveraging a neural network model of the cdf.
    \item We explore multiple techniques for improving the accuracy of both the cdf and pdf, and methods for uncertainty quantification including the use of
    conformal inference.
    \item We provide insights into modeling challenges.
\end{itemize}

Strictly speaking, Eq.\,(\ref{eq:pdf}) applies only if $x$ is from a continuous set. However, discrete distributions are frequently encountered in high-energy physics and other fields, and are often approximated by continuous distributions through suitable coarse-graining of $x$. Therefore, it is of interest to explore the degree to which we can approximate the cdf of discrete distributions with continuous ones and then apply Eq.\,(\ref{eq:pdf}) to obtain smooth approximations of the probability mass functions (pmf).

Given the pdf $f(x\mid\boldsymbol{\theta})$, where the data $D$ have been compressed to the 1-dimensional quantity $x$, further inferences can be performed by treating the pdf as a statistical model for $x$. These include constructing approximate confidence intervals by  profiling\, \cite{Algeri:2019lah} $f(x\mid \boldsymbol{\theta})$ or constructing posterior densities for the parameters $f(\boldsymbol{\theta}\mid x)$ given a suitable prior. 
However, if $x = \lambda$ is a test statistic some care must be exercised because, in general, a variation of $\boldsymbol{\theta}$ will  induce a change in $\lambda$, which should be taken into account.  
Since the quality of the estimated pdf is inherently tied to the accuracy of the cdf model, the key issue is whether a sufficiently accurate model of the cdf $F(x \mid \boldsymbol{\theta})$ can be constructed. This is the issue we explore in this paper where the focus is on test statistics that arise in simulation-based inference. 
We consider the classic ON/OFF problem of astronomy\, \cite{Li:1983fv} and high-energy physics as a benchmark example and apply the insights gained to the SIR problem \cite{kermack1927contribution, ALLEN2017128} in epidemiology. 

This paper is structured as follows. In Sec. \ref{sec:related_work} we review related work in the areas of smiulation-based inference and uncertainty quantification in neural networks. In Sec. \ref{sec:ONOFF} we model the pdf from the cdf of a test statistic for a problem where the likelihood is tractable: the prototypical signal/background problem in high-energy physics, which in astronomy is known as the ON/OFF problem. In sec. \ref{sec:SIR} we model the pdf from the cdf of a test statistic for a problem where the likelihood is intractable: the SIR model in epidemiology. In both examples, various techniques are applied to improve the accuracy of the modeled pdf and quantify the uncertainty. The paper ends with a discussion and conclusion in Sec. \ref{sec:discussion_conclusion}.

\section{Related work}
\label{sec:related_work}
\subsection{Likelihood Ratio Trick}
It has been known since at least the early 1990s \cite{80266, 80269,6795267,Prosper:1993xeg} that  probabilities can be modeled with neural networks using the density ratio method, also known as the likelihood ratio trick \cite{cranmer2016approximating, mohamed2017learning, 10.5555/3295222.3295304, goodfellow2014generative, param-NN}. Suppose one has two data samples, $\{(\mathbf{x}_i, \boldsymbol{\theta}_i) \}_A$ and $\{(\mathbf{x}_i, \boldsymbol{\theta}_i) \}_B$. Sample $A$ comprises pairs $(\mathbf{x}_i, \boldsymbol{\theta}_i)$ in which the components are sampled sequentially:
$\boldsymbol{\theta}_i \sim \pi$ followed by $\mathbf{x}_i \sim G(\boldsymbol{\theta}_i)$, where $\pi$ is a known prior and $G$ is a simulator. Sample $B$ differs from $A$ in that $\mathbf{x}_i \sim g$, where $g(\mathbf{x} \mid \boldsymbol{\theta})$ is a \emph{known} density that may, or may not, depend on $\boldsymbol{\theta}$. If sample $A$ is assigned target $t = 1$ and $B$ is assigned $t = 0$, then a sufficiently flexible function --- fitted by minimizing the cross-entropy loss averaged over a large balanced training sample --- yields an approximation to the function
\begin{align}
    D(\mathbf{x}, \boldsymbol{\theta}) &= \frac{f(\mathbf{x}, \boldsymbol{\theta})}{f(\mathbf{x}, \boldsymbol{\theta}) + g(\mathbf{x} \mid \boldsymbol{\theta}) \pi(\boldsymbol{\theta})},
\label{eq:D}
\end{align}
where $f(\mathbf{x}, \boldsymbol{\theta})$ is the (generally unknown) joint density associated with sample $A$. A rearrangement of Eq.(\ref{eq:D}) leads to the result
\begin{align}
    f(\mathbf{x} \mid  \boldsymbol{\theta}) & = \frac{f(\mathbf{x}, \boldsymbol{\theta})}{\pi(\boldsymbol{\theta})},\nonumber\\
        &= g(\mathbf{x} \mid\boldsymbol{\theta}) \left(\frac{D}{1 - D} \right).
\end{align}
This method and related methods are available in the \texttt{Madminer} package \cite{Brehmer:2019xox}.

\subsection{Flow-based Methods}
Many methods based on normalizing flows estimate a conditional pdf, for example, see \cite{conditional_NF}. Normalizing flows exploit the change of variables formula
\begin{align}
    f(\mathbf{x} \mid \boldsymbol{\theta}) & = g(z=h(\mathbf{x})| \boldsymbol{\theta}) |\det J|,
\end{align}
by modeling the function $h : X \in \mathbb{R}^d \rightarrow Z \in \mathbb{R}^d$ with a sequence of bijections, where $g(*)$ is a tractable pdf and $|\det J|$ is the determinant of the Jacobian, $J$, of the transformation.

\subsection{LF2I and ALFFI}
Our motivation for starting with the cdf (or p-value) is that constructing a model of it is the key step in the likelihood-free
frequentist (\texttt{LF2I}) approach\, \cite{dalmasso2023likelihoodfree} and in an extension  of it (\texttt{ALFFI}) \,\cite{ALFFI} in which the test statistic $x = \lambda(D; \boldsymbol{\theta})$ is included as an input to the neural network-based model of the cdf. If the test statistic $\lambda$ is chosen so that large values of $\lambda$ disfavor the hypothesis $H: \boldsymbol{\theta} = \boldsymbol{\theta}_0$ it follows that  a confidence set, $R(D)$, can be constructed at confidence level $\tau = 1 - \alpha$ where $\alpha$ is the miscoverage rate. By definition, the set $R(D)$ is all values of $\boldsymbol{\theta}_0$ for which $F(\lambda \mid \boldsymbol{\theta}_0) \le \tau$ given data $D$ \cite{dalmasso2023likelihoodfree, ALFFI}. 
The fact that $\lambda$ is an input to the model in the \texttt{ALFFI} algorithm presents an opportunity: taking the derivative of the approximate cdf $\hat{F}$ with respect to $\lambda$ provides an approximation  
$\hat{f}(\lambda \mid \boldsymbol{\theta})$ of the sampling distribution of the test statistic. In this paper we explore the accuracy with which both the cdf and pdf can be approximated. 

\subsection{Conformal Inference}
\label{sec:conf_inference}
Suppose we have $n$ training samples $\left(\mathbf{x}_i, y_i\right) \in \mathbb{R}^d \times \mathbb{R}, i \in [ 1, n ]$, 
where $x_i$ is a $d$-dimensional feature vector and $y_i$ is the response variable (target). Let $f(x)$ denote the regression function (such as a neural network), which is often fitted by minimizing the average quadratic loss between the target and the function, in which case $\hat{f}(x) \approx \mathbb{E}[y \mid x ]$. 
We are interested in predicting a new response $y_{n+1}$ from a new feature vector $x_{n+1}$. Given a miscoverage rate $\alpha \in [0,1]$ we wish to build a confidence set $\mathcal{C}_\alpha$ with the property that
\begin{equation}
\label{validity}
    \mathbb{P}\left\{y_{n+1} \in \mathcal{C}_\alpha\left(x_{n+1}\right)\right\} \geq 1-\alpha = \tau .
\end{equation}
The confidence set is a measure of uncertainty in the prediction of the function $f(x)$.

Conformal prediction (a.k.a. conformal inference) \cite{Vovk_95, Vovk_2005, Papadopoulos_2002} is a general procedure for constructing such confidence sets/intervals for any predictive model (such as a neural network). These sets are valid (i.e., they satisfy Eq. \ref{validity}) in finite samples without any assumptions about the distribution or the data other than the latter are exchangeable \cite{Vovk_95}. Split conformal prediction \cite{Lei_Wasserman_2014, Lei2016DistributionFreePI} achieves this by splitting the $n$ points into a training set and a calibration set. A regression model $\hat{f}$ is fitted on the training set and then used to predict on the calibration set. Next a conformity score, for example, $s_i = | \hat{f}(x_i) - y_i | $ is used to assess the agreement between the calibration's response variable and the predicted value. Next, define $\hat{q}_{1-\alpha}$ to be the $\lceil(n+1)(1-\alpha)\rceil / n$ quantile of the scores $s_1, ..., s_n$, where $\lceil\cdot\rceil$ is the ceiling function. Finally, for a new test feature vector $x_{n+1}$ construct the conformal interval as 
\begin{equation}
    \mathcal{C}_\alpha = [ \hat{f}(x_{n+1}) - \hat{q}_{1-\alpha},\ \hat{f}(x_{n+1}) + \hat{q}_{1-\alpha}] .
\label{eq:conformal_band}
\end{equation}
More details on conformal prediction are provided in \cite{zaffran2022adaptive, angelopoulos2021gentle}. We use conformal inference to construct confidence sets for the cdf and pdf of $\lambda$ as a way to quantify their accuracy and for potentially correcting the cdf to arrive at a more accurate cdf model.

\subsection{Multistage modeling of Neural Networks}
\label{sec:MSNN}
Recently, multi-stage neural networks (MSNN) \cite{wang2024multi} (see also \cite{aldirany2023multilevel}) has been proposed as a strategy to approximate the target function of neural networks with remarkable accuracy, with the prediction errors approaching machine precision $\mathcal{O}(10^{-16})$ for double floating point numbers.  The method constructs a sequence of neural networks, each fitted to the residuals from the previous stage. We follow the notation in \cite{wang2024multi}.

Let $x$ be the input features and $u_g(x)$ be the target function. The data $(x, u_g(x))$ are used to train a neural network $u_0(x)$ to regress $u_g(x)$. The error, or residual, between the neural network and the target function $e_1(x) = u_g(x)-u_0(x)$ is calculated. The residual data $(x, e_1(x)/\epsilon_1 ) $ are used to train a second neural network $u_1(x)$ to regress $e_1(x)/\epsilon_1 $, where $\epsilon_1$ is the root-mean square,
\begin{equation}
\label{Eq:RMS}
    \epsilon_1=\sqrt{\frac{1}{N_d} \sum_{i=1}^{N_d}\left[e_1\left(x\right)\right]^2}=\sqrt{\frac{1}{N_d} \sum_{i=1}^{N_d}\left[u_g-u_0\left(x\right)\right]^2} .
\end{equation}
The target function is rescaled so that its range is of order one. The corrected regression function is given by 
\begin{equation}
    u_c^{(1)}(x)=u_0(x)+\epsilon_1 u_1(x) .
\end{equation}

By fitting a second neural network on the residuals and adding it to the original network one improves the precision of the regression function. The algorithm can be continued to train further neural networks on $\left(x, e_n\left(x\right) / \epsilon_n\right)$ to reach higher accuracy and all $(n+1)$ neural networks are combined as follows,
\begin{equation}
    u_c^{(n)}(x)=\sum_{j=0}^n \epsilon_j u_j(x),
\end{equation}
where $\epsilon_i$ refers to the root mean square for the $i$-th neural network with $\epsilon_0=1$. We explore whether the cdf to be modeled is smooth enough for this method to work (see Sec. \ref{sec:MSNN_CDF}).

\subsection{Bootstrap Neural Networks}
\label{sec:bootstrap}
The bootstrap \cite{Efron_79} is a statistical method that treats an observed dataset as if it were a population. This makes it possible, for example, to approximate the sampling distributions of statistics.
Given an estimator $\hat{\theta}$ of $\theta$, the bootstrap quantifies the uncertainty in $\hat{\theta}$ by repeatedly sampling the original dataset with replacement to create new datasets. The uncertainty in $\hat{\theta}$ can be quantified with a measure of the variability of predictions between different bootstrap datasets \cite{bishop:2006:PRML, clarte2024analysis}.

In the context of supervised learning, the samples are given in pairs $z_i = (\mathbf{x}_i, y_i)$ from a joint distribution $p_\theta(\mathbf{x},y)$. Suppose we are given a set of features and targets composed of $n$ examples $\mathbf{z} = \{ \left(\mathbf{x}_i, y_i\right)\}_{i=1}^n$. One draws a bootstrap dataset from the original training dataset of the same size as the original dataset. 
This is done $K$ times and the same model is fitted to each of the $K$ bootstrap datasets using the same training protocol. This yields $K$ neural networks and, therefore, $K$ outputs for a given input. A measure of the spread of the outputs quantifies the uncertainty in the model output. 

\section{Example 1: signal/background or ON/OFF model}
\label{sec:ONOFF}
For our first example, we choose a problem that is ubiquitous in high-energy physics and astronomy: the signal/background problem, also known as the ON/OFF problem in astronomy \cite{Li:1983fv} and we consider its simplest realization. An observation is made which consists of counting $N$ events (these are particle collisions in particle physics or photon counts in astronomy). A second independent observation is made where no signal is present by design, yielding $M$ counts.  Following \cite{ALFFI} the statistical model is taken be a product of two Poisson probability mass functions (pmf) for the observed data $D = \{N, M \}$,
\begin{align}
    \mathbb{P} (N, M \mid \mu, \nu) = \mathcal{L} ( D; \mu, \nu) & = \frac{(\mu + \nu)^N \exp(-(\mu + \nu))}{N!} \,\frac{\nu^M\exp(-\nu)}{M!},
\label{Eq:OnOffModel}
\end{align}
where $\mu$ and $\nu$ are the mean signal and background counts, respectively. The same likelihood ratio test statistic is used as in \cite{ALFFI}
\begin{equation}
    \lambda(D ; \mu, \nu)  = -2 \log \left[  \frac{\mathcal{L}(D;\mu, \nu)}{\mathcal{L}( D; \hat{\mu}, \hat{\nu})} \right],
    \label{Eq:lambda_on_off}
\end{equation}
where the maximum likelihood estimate of the signal is $\hat{\mu} = N - M$, which can be positive or negative. Negative signal estimates are avoided by using the ``non-maximum likelihood estimate'' 
\begin{equation}
     \hat{\mu} =\left\{
    \begin{array}{ll}
        N-M & \text{ if } \quad  N>M \\
        0 & \quad \textrm{ otherwise.}
    \end{array}
    \right. 
    \label{eq:muhat}
\end{equation}
and 
\begin{equation}
     \hat{\nu} =\left\{
    \begin{array}{ll}
        M & \text{ if } \quad  \hat{\mu} = N - M \\
        (M+N)/2 & \quad \textrm{ otherwise.}
    \end{array}
    \right. 
    \label{eq:nuhat}
\end{equation}
Since $\mu$ is the \emph{parameter of interest} and $\nu$ is a \emph{nuisance parameter} one is generally interested in confidence intervals for $\mu$ regardless of the true value of $\nu$. However, the sampling distribution of $\lambda$, $f(\lambda \mid \mu,\nu)$, in general depends on both the parameter of interest and the nuisance parameter. This remains true even if we were to replace
Eq. \ref{Eq:lambda_on_off} with the profile likelihood ratio in which the nuisance parameter $\nu$ is replaced by its conditional estimate. This is because of the exclusion of negative estimates of the signal. Therefore, we still must contend with the nuisance parameter.

There are at least two plausible ways one might proceed given an approximation to the sampling distribution, $f(\lambda \mid \mu,\nu)$: either replace the nuisance parameter by a plug-in estimate or replace it through a procedure analogous to profiling but applied to $f(\lambda \mid \mu,\nu)$. However, while profiling one has to account for the fact that $\lambda(D; \mu, \nu)$ depends on the nuisance parameter.  The viability of using a profiled sampling distribution for creating approximate confidence intervals for $\mu$ alone remains to be explored.



\subsection{Modeling the cdf with ALFFI}
\texttt{ALFFI} is an algorithm for approximating the conditional cdf, $F(x \mid \theta)$, of a scalar random variable $x$ where $\theta$ denotes the parameters of the statistical model. If $x = \lambda(D; \theta)$ is a test statistic then the cdf can be used to construct confidence sets for all parameters simultaneously. Here we explore the accuracy of the \texttt{ALFFI} algorithm for modeling the cdf with a view to deriving the sampling distribution, $f(\lambda \mid \mu, \nu)$ of the ON/OFF test statistic by differentiating the cdf with respect to $\lambda$.  

Again following \cite{ALFFI} we start by sampling $\mu$ and $\nu$ from uniform priors and sample  $n_i \sim \text{Poiss}(\mu_i+\nu_i)$
and $m_i \sim \text{Poiss}(\nu_i)$. At each parameter point, we calculate $\lambda_i = \lambda (n_i, m_i \mid \mu_i, \nu_i)$ according to Eq. \ref{Eq:lambda_on_off} and the procedure is repeated to sample $N_i, M_i$ and $\lambda_D =\lambda (N_i, M_i \mid \mu_i, \nu_i) $. Finally, the indicator $Z$, which is unity if $\lambda_i \le \lambda_D$ and zero otherwise, is computed. This results in a training set of size $B$, $\mathcal{T} = \{ (\mu_i, \nu_i, \lambda_i, Z_i) \}_{i=1}^B$. The key observation in 
\texttt{ALFFI} 
and \texttt{LF2I} is that critical value functions, such as the cdf $F(\lambda \mid \mu, \nu)  = \mathbb{P}( \lambda \le \lambda_D \mid \mu, \nu)$, are the expectation value $\mathbb{E}(Z \mid \lambda_D, \mu, \nu)$ of the discrete random variable $Z$, of which a smooth approximation can be created with a deep neural network $f(\mathbf{x}_i, \mathbf{w})$. Critically, this network must be trained to minimize the mean square error (MSE) loss\footnote{In \texttt{LF2I} and \texttt{ALFFI}, the predicted value is $\mathbb{E}[Z \mid x] = \mathbb{P}(Z=1 \mid x) = \frac{\mathbb{P}( x \mid Z=1) \mathbb{P}(Z=1) }{\mathbb{P}( x \mid Z=1) + \mathbb{P}( x \mid Z=0)}$. In other words, it behaves like a classifier, and because of that the cross entropy loss can also be used.}, 
\begin{equation}
\label{MSE_loss}
    L(\mathbf{w}) = \frac{1}{B} \sum_{i=1}^B \left( y_i -f(\mathbf{x}_i, \mathbf{w}) \right)^2,
\end{equation}
where $\mathbf{w}$ are the parameters (weights) of the neural network and $\mathbf{x}_i = \{ \mu_i, \nu_i, \lambda_i \}$ is a batch of training data, with targets $y_i=Z_i$. Details of this algorithm are provided in Appendix \ref{ALFFI_alg_OnOff}.

The cdf, using the \texttt{ALFFI} algorithm, was approximated using a fully-connected neural network with 3 input features $\mathbf{x} = \{ \mu, \nu, \lambda \}$, 12 hidden layers with 12 nodes each, and a single output. A sigmoid was used in the output layer to constrain the output to the unit interval. The activation function at each hidden node is a SiLU \cite{SiLU}. The SiLU nonlinearity was used in all the models in this example, since the SiLU allows the model to be smooth and differentiable multiple times. 
The network was trained with the NAdam optimizer \cite{NAdam} with a fixed learning rate of $6 \times 10^{-4}$. The training set is composed of $10^7$ examples, which were used in batches of size $512$ for  $10^5$ iterations, that is, for 5 epochs. Longer training runs were also performed, but did not yield improvements in the results.

\begin{figure*}[ht]
  \centering
  \begin{minipage}[b]{\textwidth}
    \centering
    \includegraphics[width=\textwidth,height=100mm]{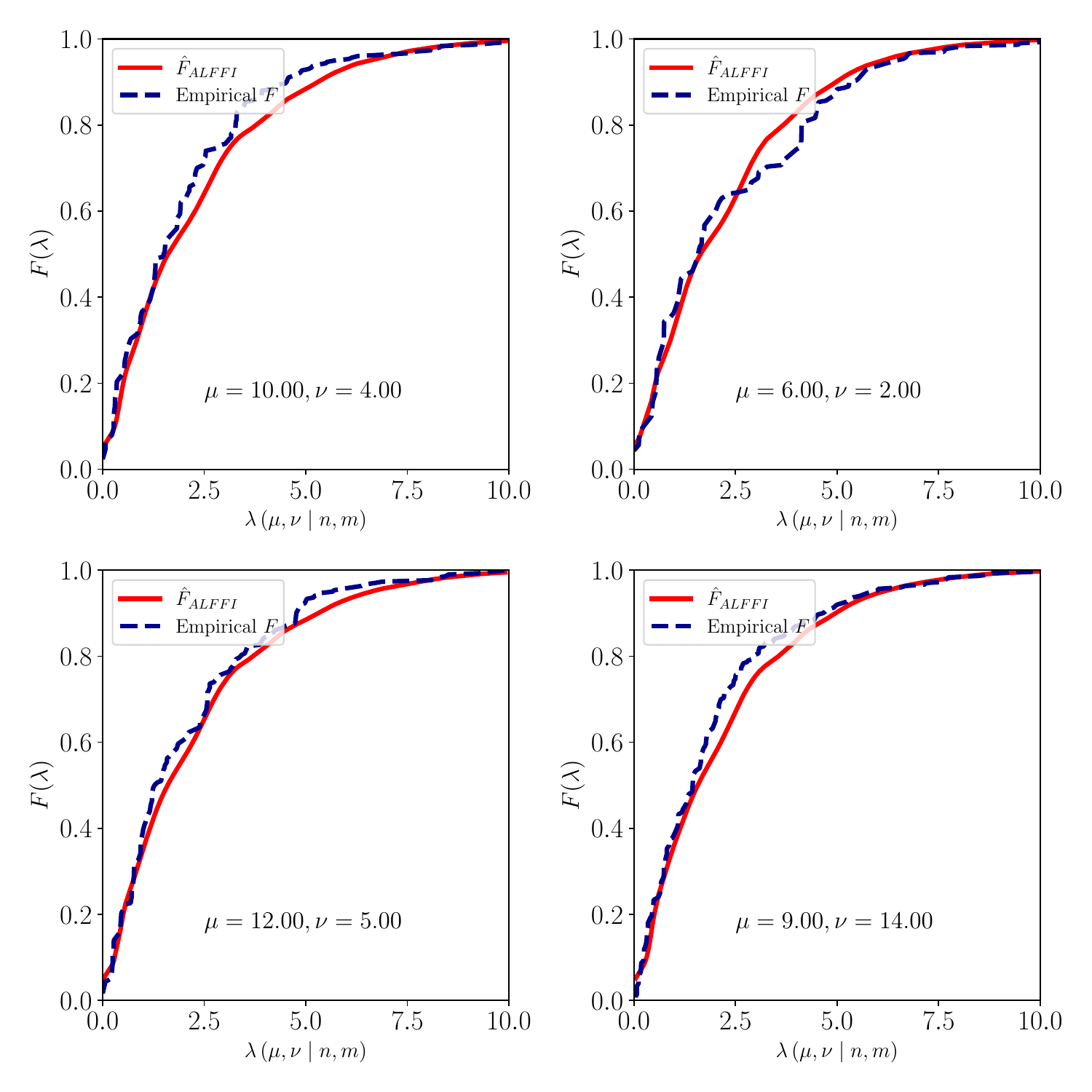}
    \caption{ON/OFF problem: cdfs modeled with \texttt{ALFFI} at different $(\mu,\nu)$ points.}
    \label{ALFFI_CDF}
  \end{minipage}
\end{figure*}

\begin{figure*}[ht]  
  \begin{minipage}[b]{\textwidth}
    \centering
    \includegraphics[width=\textwidth,height=100mm]{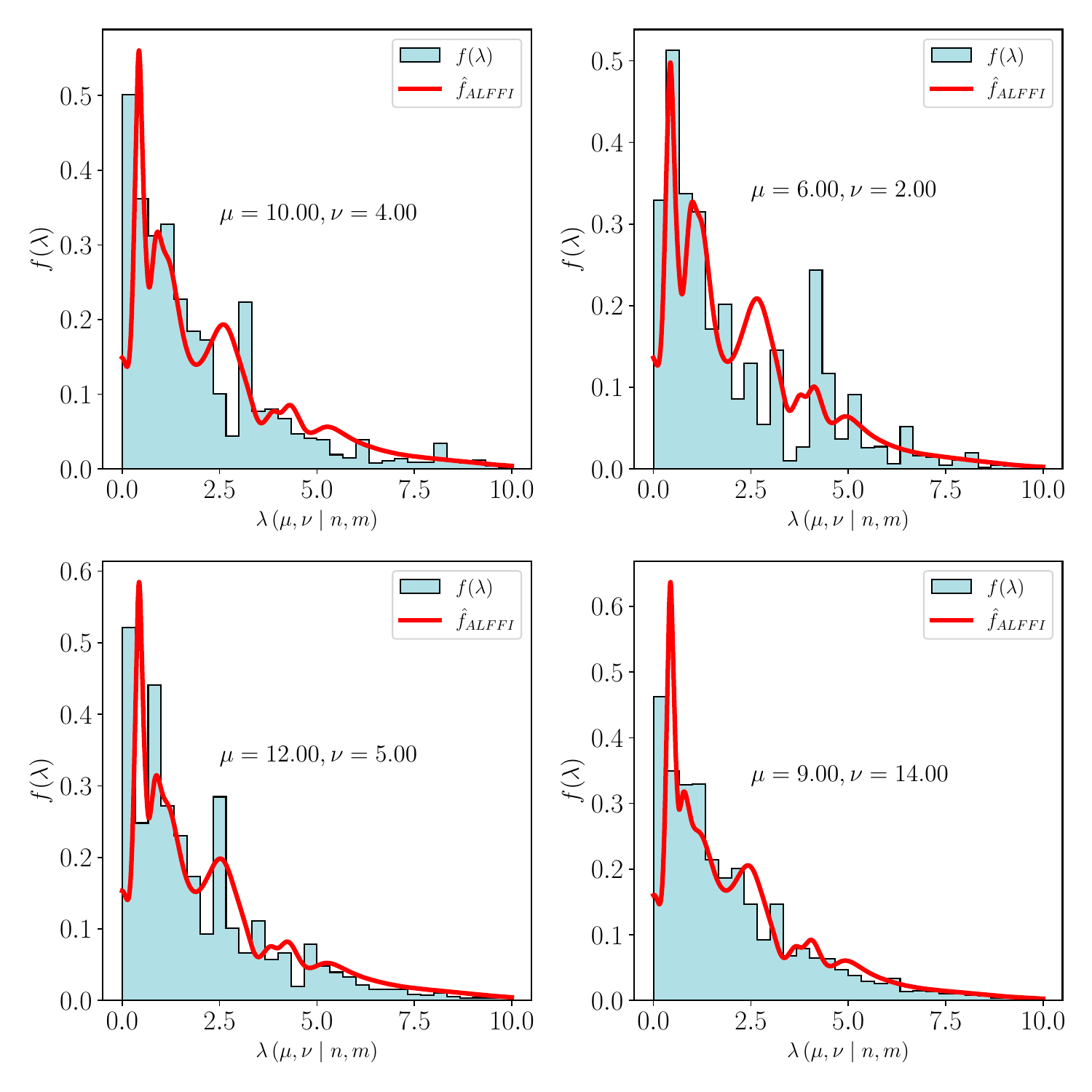}
    \caption{ON/OFF problem: pdfs obtained by differentiating cdfs modeled with ALFFI at different $(\mu,\nu)$ points.}
    \label{ALFFI_PDF}
  \end{minipage}
\end{figure*}

Although the \texttt{ALFFI} algorithm yields valid confidence sets for the ON/OFF problem, it does not yield a sufficiently accurate approximation of the cdf, and hence of the pdf. This is shown in Fig.\,\ref{ALFFI_CDF}, where the empirical cdf\footnote{Suppose that we have a sample of observations $\lambda_1, ..., \lambda_n$. The empirical cdf is defined by $F(\lambda)=\frac{1}{n} \sum_{i=1}^n \mathbb{I}\left( \lambda_i \leq \lambda \right)$}, $F$, is displayed together with the approximated cdf  $\hat{F}_{\text{ALFFI}}(\lambda\mid \mu, \nu)$. 
Taking the derivative of the approximated cdf yields the \texttt{ALFFI} approximation of the pdf, $\hat{f}_{\text{ALFFI}}(\lambda)=\frac{\partial \hat{F}_{\text{ALFFI}}}{\partial \lambda}$, displayed in Fig. \ref{ALFFI_PDF}. We observe that $\hat{f}_{\text{ALFFI}}(\lambda)$ displays sharp fluctuations and turning points, which was also confirmed by taking a numerical derivative of $\hat{F}_{\text{ALFFI}}$. This is because the actual slope of the cdf changes dramatically from point to nearby point, especially at low values of $\lambda$. It can also be seen that the histogrammed pmf of $\lambda$ at a particular $(\mu, \nu)$ point is a high frequency function, or one that is noisy and features very sharp peaks. The fluctuations seen in the pmf also depend on the choice of binning: finer binning shows larger fluctuations because this is a discrete distribution. This high-frequency nature of the pmf  complicates our attempt to create a smooth modeling of the pmf with \texttt{ALFFI}.

\subsection{Directly Modeling the Empirical CDF}

\begin{figure*}[h]
\label{chi2_conv}
  \centering
  \begin{minipage}[b]{\textwidth}
    \centering
    \includegraphics[width=\textwidth,height=100mm]{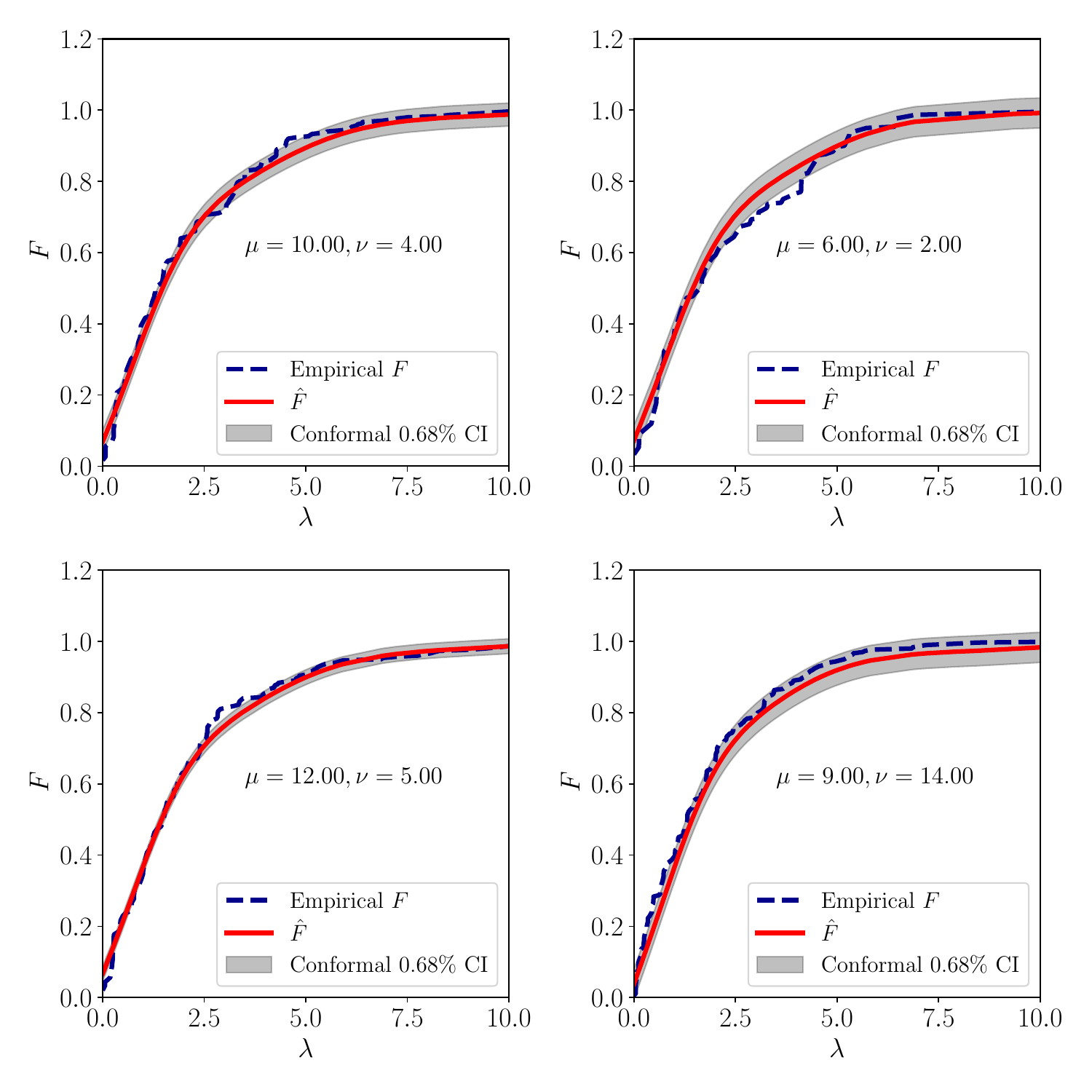}
    \caption{ON/OFF problem: cdfs from modeling the empirical cdf with the latter as the targets at different $(\mu,\nu)$ points, with the associated $68\%$ confidence band computed using conformal inference.}
    \label{EmpiricalCDF_CDF}
  \end{minipage}
\end{figure*}

\begin{figure*}[h]
  \begin{minipage}[b]{\textwidth}
    \centering
    \includegraphics[width=\textwidth,height=100mm]{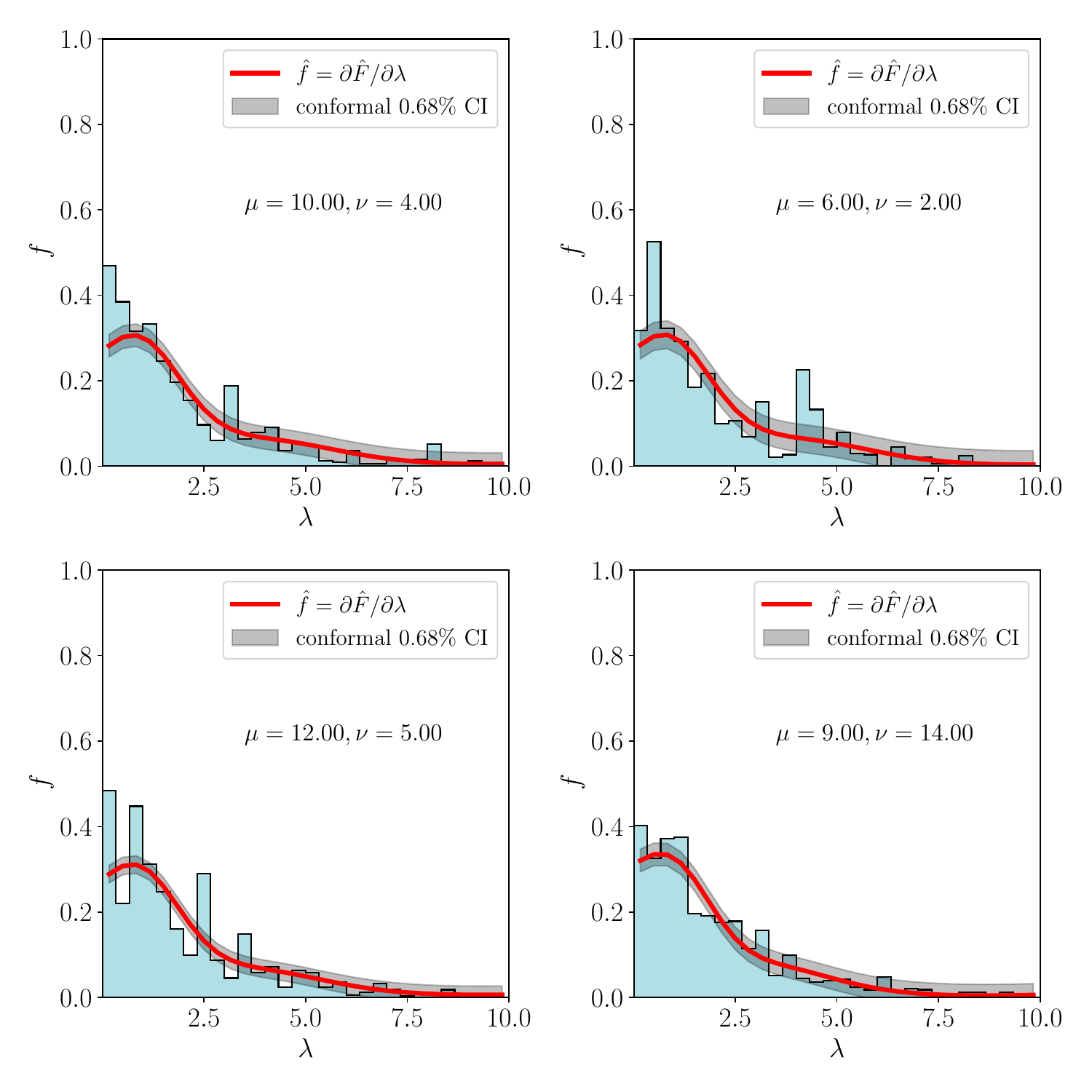}
    \caption{ON/OFF problem: pdfs obtained by differentiating the models of the empirical cdf at different $(\mu,\nu)$ points, with the associated $68\%$ confidence band computed using conformal inference.}
    \label{EmpiricalCDF_PDF}
  \end{minipage}
\end{figure*}

Trying to model an intrinsically discrete distribution with a smooth approximation may be beyond an algorithm such as \texttt{ALFFI} and one might argue it is in any case a hopeless task. Nevertheless it is useful to check if an alternative, more direct, approach works better: directly modeling the empirical cdf. At each $\{ \mu_i, \nu_i \}$ point, we generate $K=100$ experiments:  $\{N_i^{(j)} \}_{j=1}^K \sim \text{Poiss}(\mu_i+\nu_i)$ and $\{ M_i^{(j)} \}_{j=1}^K \sim \text{Poiss}(\nu_i)$, for which we generate $K$ instances of the test statistic $\{\lambda_i^{(j)}\}_{j=1}^K$ according to Eq. \ref{Eq:lambda_on_off}. These data are used to calculate the empirical cdf $F_i(\lambda_i)$ at every parameter point. The data were then flattened to result in a training set of size $B=10^7$, $\mathcal{T} = \{ (\mu_i, \nu_i, \lambda_i, F_i) \}_{i=1}^B$. 

The cdf was again approximated with a NN with 3 input features $\mathbf{x} = \{ \mu, \nu, \lambda \}$ but this time with the targets set to values of the empirical cdf $F$. The architecture consisted of 6 hidden layers with 12 nodes each and a single output. The activation function at each hidden node is a SiLU. The network was trained to minimize the MSE loss with the NAdam optimizer with a fixed learning rate of $3\times 10^{-5}$. The training set is composed of $10^7$ examples, which were used in batches of size $60$ for $10^6$ iterations, that is, for 6 epochs.

A much better approximation of the cdf is obtained, as shown in Fig. \ref{EmpiricalCDF_CDF} and more importantly, the associated pdfs in Fig. \ref{EmpiricalCDF_PDF} show that the jagged fluctuations in the pdf disappear and we obtain a smooth approximation of the discrete distribution. Modeling the empirical cdf presents yet another advantage: the possibility of using the empirical cdf as the true response variable in the conformal inference algorithm (see Sec. \ref{sec:conf_inference}) and thereby providing a quantification of the uncertainty in the modeling of the cdf and pdf. We use conformal inference to calculate a $68 \%$ confidence interval in the cdf space for each value of the test statistic. For the associated pdf, a coarse-graining of $\lambda$ is performed by histogramming $\lambda$ and the coarse-grained $f(\lambda)$ was used as the true response variable (that is, the true pdf). The excellent agreement between the predicted pdf and the histogrammed pdf is shown in Fig. \ref{EmpiricalCDF_PDF} along with the $68 \%$ conformal confidence interval at each value of $\lambda$.

\subsection{Multistage Modeling of the CDF}
\label{sec:MSNN_CDF}
Another approach to improve the accuracy of regression functions is described in \cite{wang2024multi}, which proposes a multi-stage neural network (MSNN) strategy in which the network training is divided into multiple stages, where each stage fits a separate neural network to the residuals from the previous stage (see Sec. \ref{sec:MSNN}).  

 \textbf{Quantile residuals ($\Delta C$):}  
Let $p=\{F_1,..., F_k \}$ be a set of evenly spaced probabilities and let $\hat{F} =\{ \hat{F}_1, ..., \hat{F}_n \}$ be a set of estimated cdf values at a given $(\mu,\nu)$ parameter point with $n \gg k$. Define $q$ as the quantiles of $\hat{F}$ associated with the probabilities $p$. If we can model the correction function $p = C(q)$ that maps the estimated cdfs $q$ to the exact cdf $p$ with sufficient accuracy then, in principle, the correction function can be used to improve the estimated cdf. Writing $p = q + \Delta C(q)$,  we define the quantile residual function as $\Delta C = q-p$ and try to model it as a function of $q$ following the multistage method.

   In order to follow the multistage method for constructing a model of the cdf one needs an accurate model of the residuals. We anticipate that this will be  challenging because the distribution of $\lambda$ is discrete and their values are not evenly spaced. Indeed, this is borne out in Fig. \ref{fig:deltaC} which shows that $\Delta C$ is a noisy high frequency function. Furthermore, as can be seen in Fig. \ref{fig:deltaC}, the functions do not seem to vary smoothly with the parameters $(\mu,\nu)$ (small changes in the parameter values yield large changes in the residual function).

Nevertheless, we wish to see whether a neural network can produce a smooth interpolation of this noisy function by implementing the suggestions described in \cite{wang2024multi}, which are: 
\begin{enumerate}
    \item Use a sine nonlinearity in the first layer and tanh nonlinearity in subsequent layers.
    \item Use Glorot weight initialization \cite{Glorot} with a constant factor $\kappa=60$ multiplying the weights in the first layer and no such factor in subsequent layers. It was observed that the $\kappa$ weight factor helps to stabilize the training and the sine nonlinearity helps capture the high frequency.
    \item Normalize the target function by the RMS of the outputs of the target function $ \Delta C \rightarrow \Delta C/RMS( \Delta C)$, as in Eq. \ref{Eq:RMS}.
\end{enumerate}

Further experiments were performed such as adding or omitting dropout, batch normalization, weight regularization, and experimenting with different nonlinear functions such as ReLU, SiLU, tanh, sine, and groupsort \cite{groupsort}. None of these experiments were able to capture the high frequency behavior of the residual target function when trained on the residuals of all parameters in the training data simultaneously. However, we were able to use a simple neural network to model the residual function when modeled point-by-point in the parameter space.

The frequency principle \cite{xu2019frequency} states that neural networks first fit low frequencies and later fit high frequencies during the training process. However, prior studies \cite{tancik2020fourier, wang2021eigenvector} have demonstrated that in practice standard NNs struggle to fit high frequency functions. We confirm these results and observe that the NNs we used struggle to fit the residual functions even when trained for very long training runs. LSTMs, \cite{LSTM} on the other hand, are known to outperform standard NNs in fitting high frequency and sequential data, where the order of the data is significant \cite{LINDEMANN2021650}. They have been successfully applied in many areas in sequence learning and time series forecasting such as financial forecasting \cite{BHANDARI2022100320}, handwriting recognition \cite{LSTM_handwriting} and speech recognition \cite{LSTM_speech_recognition}. In our study, LSTMs achieve reasonably good predictions of the residuals when trained over all $(\mu, \nu)$ values. 
Although our LSTM models outperform our multi-layer perceptron (MLP) models in this task, the results are unsuitable for use in the MSNN approach due to inadequate precision in the resulting predictions.

\textbf{Cdf residuals ($\Delta r$):} An alternative to modeling the quantile residuals is to model the cdf residuals as a function of $\lambda$, which is closer to what is done in the MSNN approach. At a particular $(\mu, \nu)$ point, the cdf residual is simply the difference between the predicted and the empirical cdf: $\Delta r = \hat{F}-F$. Figure\,\ref{fig:deltar} shows that the cdf residuals as a function of $\lambda$ are a slightly smoother function of the parameter points than is $\Delta C$. Nevertheless, the MLP and LSTM models still struggle to capture the high-frequency nature of this residual function. A coarse-graining of $\lambda$ was also performed to reduced the high-frequency of the residual function but to no avail.

\section{Example 2: SIR Model}
\label{sec:SIR}
The lessons learned in the ON/OFF example are applied to modeling the cdf in a different context: epidemiology. We use the SIR model \cite{kermack1927contribution, ALLEN2017128} which treats an epidemic as a system in which transitions occur between three states or compartments: susceptible (S), infected (I), and recovered (R). This model has both a stochastic realization as well as a description in terms of ordinary differential equations 
\begin{align}
    \frac{dS}{dt} & = - \beta \, S I, \nonumber\\
    \frac{dI}{dt} & = - \alpha I + \beta \, S I,\nonumber\\
    \frac{R}{dt} & = \alpha I ,
    \label{eq:SIR}
\end{align}
whose solutions approximate the mean number of susceptible $(S(t))$, infected ($I(t)$) and recovered ($R(t)$) individuals as a function of time, $t$. The model depends on two parameters $\alpha$, the rate of recovery, and $\beta$, the rate of transmission per infected individual. We fit the model to the data $D=\{ x_1, ..., x_n \}$ by minimizing the following test statistic \cite{ALFFI}
\begin{align}
    \lambda(D; \theta) & = \frac{1}{50} \sqrt{\frac{1}{N} \sum_{n=1}^N \frac{[x_n - I_n]^2}{I_n}},
\label{eq:lambda_SIR}
\end{align}
where $\theta = \{\alpha, \beta \}$, $x_n$ is the observed number of infected individuals at the observation time $t_n$ and $I_n = I(t_n, \theta)$ is the predicted mean infected count obtained by solving the system of ordinary differential equations, Eq. \ref{eq:SIR}.
We have chosen this example because it demonstrates the practical utility of the methods discussed in this paper for performing inference with intractable statistical models. While all of our major findings from the ON/OFF example apply here, our focus in this section is to improve the modeling of the cdf and to employ additional techniques for uncertainty quantification.


\subsection{Directly Modeling the Empirical CDF and Sensitivity Analysis}
\label{Direct_CDF_SIR}
One of the lessons from the ON/OFF problem is that for discrete distributions the \texttt{ALFFI} algorithm does not yield a sufficiently accurate smooth model of the cdf, which was confirmed in the SIR example. We therefore directly model the empirical cdf as a function of the model parameters $(\alpha, \beta)$ and the test statistic. The training dataset comprises $250$ uniformly sampled $(\alpha, \beta)$ parameter points, where at each $(\alpha_i, \beta_i)$ point, $K=400$ epidemics were simulated yielding a set of test statistics $\{ \lambda_i^{(j)} \}_{j=1}^K$ and a set of empirical cdfs $\{ F_i^{(j)} \}_{j=1}^K$. The data were then flattened to result in a training set of size $B \times K=10^5$, $\mathcal{T} =  \{\alpha_{(i)}, \beta_{(i)}, \lambda_{(i)}, F_{(i)}(\lambda_{(i)}) \}_{i=1}^{B \times K}$. All of the models for this example approximate the cdf using a fully-connected neural network with 3 input features $\mathbf{x} = \{ \alpha, \beta, \lambda \}$ and the targets the values of the empirical cdfs $F$.

To optimize predictive accuracy, we use the \texttt{Optuna} \cite{optuna} framework to fine-tune both the neural network architecture and its hyperparameters. 
\texttt{Optuna} uses a Tree-Structured Parzen Estimator (TPE) \cite{TPE} algorithm, which is a Bayesian optimization method specifically designed for hyperparameter optimization in machine learning models.
The  hyperparameter search space is specified in Table \ref{tab:hyperparameters}. 40 optimization trials were executed, where each trial involved sampling a unique neural network configuration and training regimen from the defined search space. For every sampled set of hyperparameters, a neural network was instantiated and trained for 5 epochs. Upon completion of each trial, the model's performance was evaluated on a validation set, and the best MSE loss achieved was recorded as the objective value.

\begin{algorithm}[H]
\caption{SIR example: data generation, hyperparameter optimization, training and inference}
\label{alg:SIR_algorithm}
 \hspace*{\algorithmicindent} \textbf{Input:} number of parameter points $B$; number of simulated epidemics per point $K$; observed number of infected individuals $D=\{ x_1, ..., x_n \}$; number of hyperparameter optimization trials $N$; number of training iterations $M$; hyperparameter search space $\Phi$; miscoverage rate $\alpha \in [0,1]$ \\
 \hspace*{\algorithmicindent} \textbf{Output:} learned pdf $\hat{f}(\lambda \mid \alpha, \beta)$ with $1-\alpha$ conformal inference band
\begin{algorithmic}[1]
\State \textbf{// data generation}
\State Set $\mathcal{T'} \gets \emptyset$
\For{$i$ in $\{1,...,B \}$}
    \State Draw $\alpha_i \sim \pi_{\alpha}$, $\beta_i \sim \pi_{\beta}$
    \For{$j$ in $\{1,...,K \}$}
    \State $I_i^{(j)}$ = SimulateEpidemic($\alpha_i, \beta_i$) as in Eq. \ref{eq:SIR}
    \State Compute test statistic $\lambda_i^{(j)} = \lambda (D; \alpha_i, \beta_i, I_i^{(j)})$ as in Eq. \ref{eq:lambda_SIR}
    \EndFor
\State Compute empirical cdf $F_i(\lambda) = F (\lambda | \lambda_i^{(1)}, ..., \lambda_i^{(K)} )$   
\EndFor
\State Flatten $\mathcal{T} \gets \mathcal{T} = \{\alpha_{(i)}, \beta_{(i)}, \lambda_{(i)}, F_{(i)}(\lambda_{(i)}) \}_{i=1}^{B \times K}$
\State \textbf{// hyperparameter optimization}
\For{$j$ in $\{1,...,N \}$}
    \State Initialize a candidate vector of hyperparameters $\vec{\phi}_j \in \Phi$ and NN model $\mathcal{M}_{\vec{\phi}_j}$
    \State Train $\mathcal{M}_{\vec{\phi}_j}$ to regress $F$ on $\{ \alpha, \beta, \lambda \}$ for 5 epochs
    \State Compute final loss $c_j = \mathcal{L}(\mathcal{M}_{\vec{\phi}_j}, \mathcal{D}_{\text{train}}, \mathcal{D}_{\text{valid}})$
\EndFor
\State Find the optimal vector of hyperparameters $\vec{\phi}^*$ that corresponds to the lowest validation loss $\{ c_j \}_{j=1}^N$ 
\State \textbf{// training}
\State Initialize model with optimal vector of hyperparameters $\mathcal{M}_{\vec{\phi^*}}$
\State Use $\mathcal{T}$ to learn cdf $\hat{F}(\lambda \mid \alpha, \beta)$ via regression of $F$ on $\{ \alpha, \beta, \lambda \}$ for $M$ iterations using model $\mathcal{M}_{\vec{\phi^*}}$
\State \textbf{// inference}
\State Compute approximated pdf $ \hat{f}(\lambda \mid \alpha, \beta) = \frac{\partial \hat{F}(\lambda \mid \alpha, \beta)}{\partial \lambda}$ with autograd
\State Compute $1-\alpha$ conformal inference confidence band as in Eq. \ref{eq:conformal_band}
\State \Return learned pdf $\hat{f}(\lambda \mid \alpha, \beta)$ with $1-\alpha$ conformal inference band
\end{algorithmic}
\end{algorithm}

\begin{figure}[h]
    \centering
    \begin{minipage}{\textwidth}
        \centering
        \includegraphics[width=\linewidth]{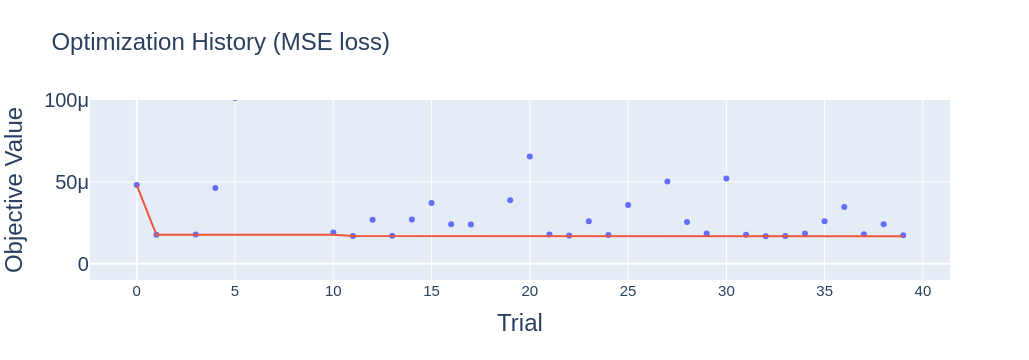} 
    \end{minipage}\hfill
    \begin{minipage}{\textwidth}
        \centering
        \includegraphics[width=\linewidth]{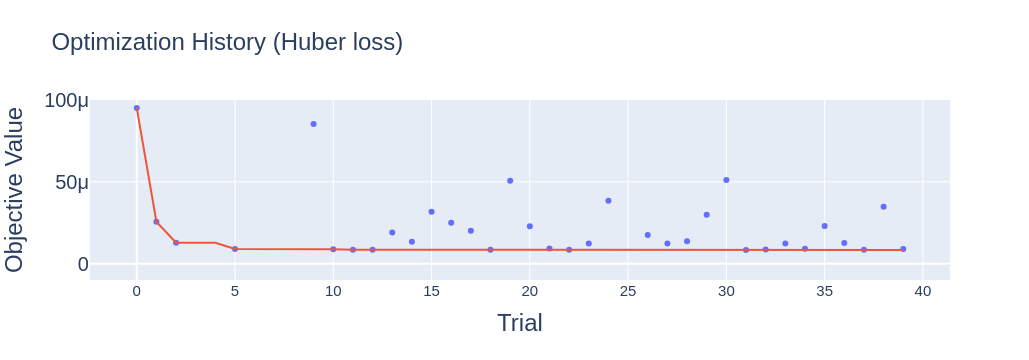} 
    \end{minipage}
\caption{Optimization history using MSE loss (top) and using Huber loss (bottom). For each trial, the blue dots represent the objective value. The red line shows the best objective value obtained up to a given trial.}
\label{optimization_history}
\end{figure}

\begin{figure}[h]
    \centering
    \begin{minipage}{\textwidth}
        \centering
        \includegraphics[width=\linewidth]{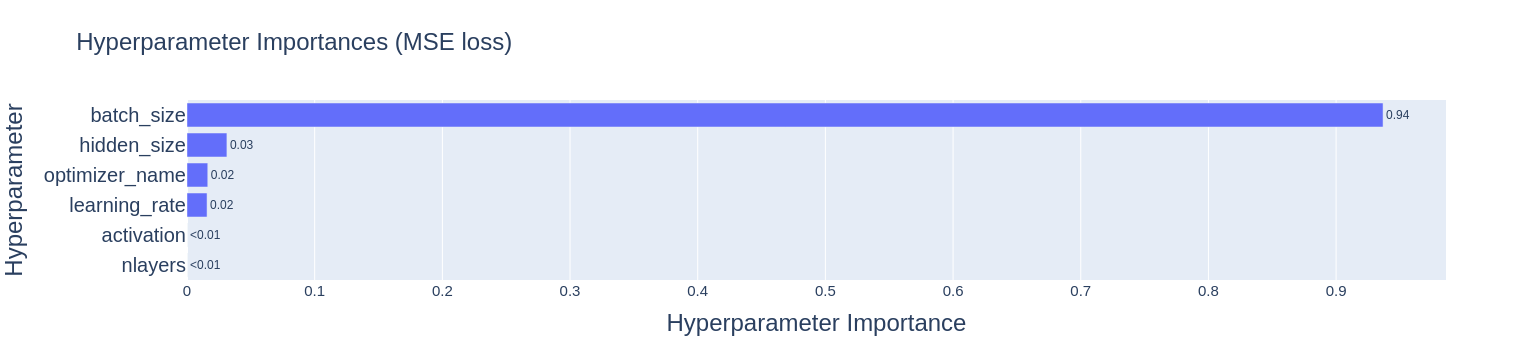} 
    \end{minipage}\hfill
    \begin{minipage}{\textwidth}
        \centering
        \includegraphics[width=\linewidth]{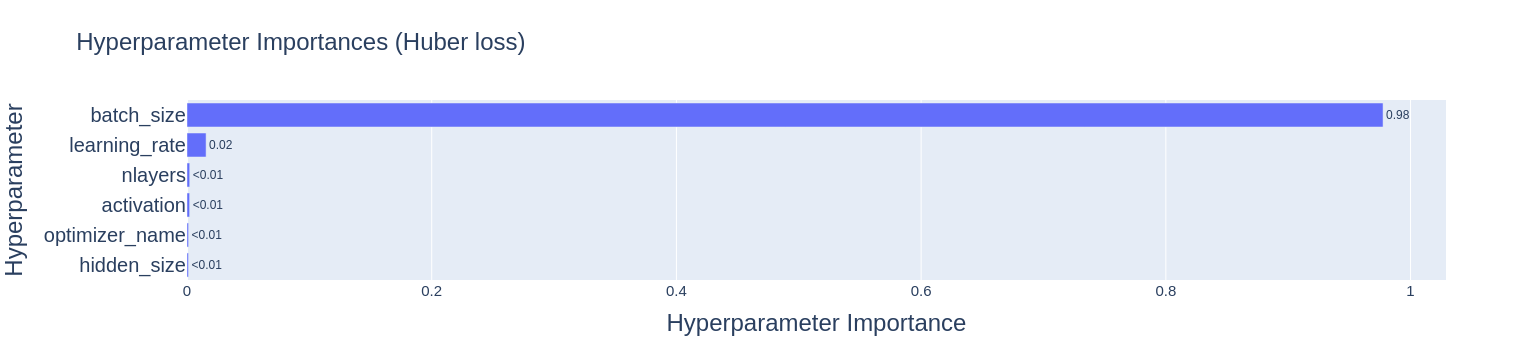} 
    \end{minipage}
\caption{Hyperparameter importances using MSE loss (top) and using Huber loss (bottom).}
\label{hyperparaneter_importances}
\end{figure}

The MSE loss is known to be sensitive to outliers in the data. Therefore, we  also used the more robust Huber loss \cite{Huber_64}, 
\begin{equation}
L_\delta(\mathbf{w}) = \left\{ 
\begin{array}{ll}
\frac{1}{2}(y-f(\mathbf{x},\mathbf{w}))^2 & \text{for } |y-f(\mathbf{x},\mathbf{w})| \leq \delta \\
\delta \left(|y-f(\mathbf{x},\mathbf{w})| - \frac{1}{2} \delta\right), & \text{otherwise,}
\end{array} 
\right. 
\end{equation}
where $\delta$ is a tunable parameter, chosen to be $0.7$. The Huber loss, which is differentiable everywhere and robust to outliers, is an MSE loss for relatively small errors and an absolute loss for larger errors.  We compare the performances of the two losses  by repeating the hyperparameter optimization procedure outlined above  using the Huber loss. 
The resulting best hyperparameter values for both losses are reported in Table \ref{tab:hyperparameters} and the hyperparameter importances are shown in Fig. \ref{hyperparaneter_importances}. The resulting optimization history for both losses is reported in Fig. \ref{optimization_history}, which shows that the optimization is robust with respect to the choice of hyperparameters and network architecture.

Our analysis reveals that the mini-batch size is the most influential hyperparameter affecting both loss functions, with larger batch sizes consistently leading to enhanced performance. This observation indicates that, for this particular problem, reducing the stochastic noise in gradient estimations—achieved by increasing the mini-batch size—yields better optimization results. Consequently, using larger mini-batches appears to be beneficial, as it provides more accurate gradient calculations and promotes more stable convergence during training.

\begin{table}[h]
    \centering
    \begin{tabular}{|l|l|l|l|}
    \hline
    \textbf{Parameter} & \textbf{Search space} & \textbf{MSE} & \textbf{Huber} \\ \hline
    Number of layers & $\in [1,6]$ & $3$ & $2$ \\ \hline
    Number of neurons & $\in [1,64]$ & $55$ & $60$ \\ \hline
    Optimizer & [Adam, NAdam, RMSprop, SGD] & RMSprop &Adam \\ \hline
    Learning rate & $\in [10^{-6}, 10^{-2}]$ &$0.00268$ & $0.00543$ \\ \hline
    Batch size & $\in [50,3\times 10^4 ]$ & $19,980$ & $19,961$ \\ \hline
    Activation & [ReLU, LeakyReLU, SELU, PReLU] & LeakyReLU &PReLU \\ \hline
    \end{tabular}
    \caption{Hyperparameter tuning ranges and best values. The right two columns show the best values from the optimization process for the MSE and Huber losses, respectively.}
    \label{tab:hyperparameters}
\end{table}

\begin{figure*}[h]
    \centering
\includegraphics[width=\textwidth,height=70mm]{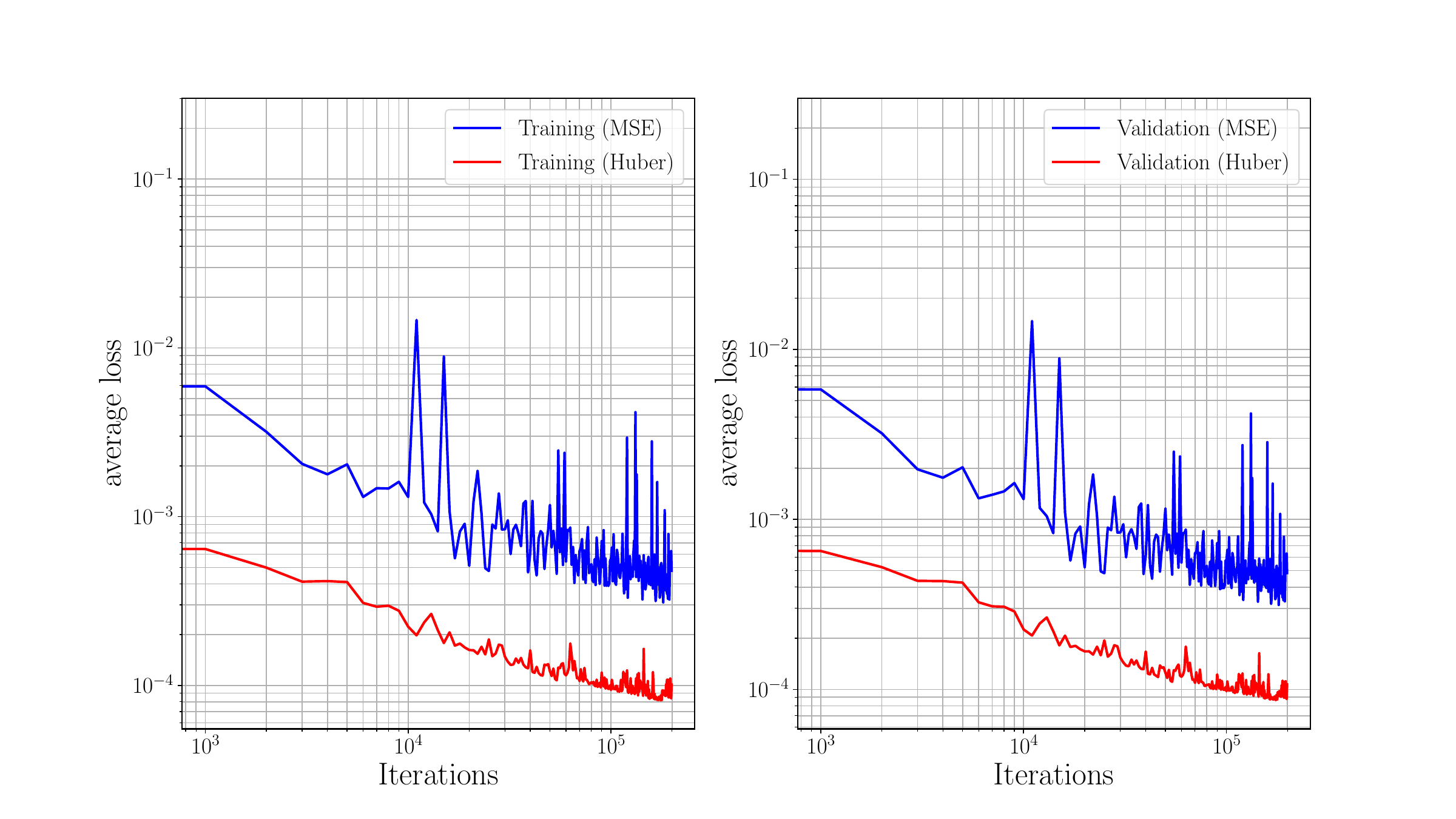}
    \caption{SIR problem: training (left) and validation (right) loss curves.}
  \label{fig:loss_curves}
\end{figure*}

\begin{figure*}[h]
  \centering
  \begin{minipage}[b]{\textwidth}
    \centering
    \includegraphics[width=\textwidth,height=100mm]{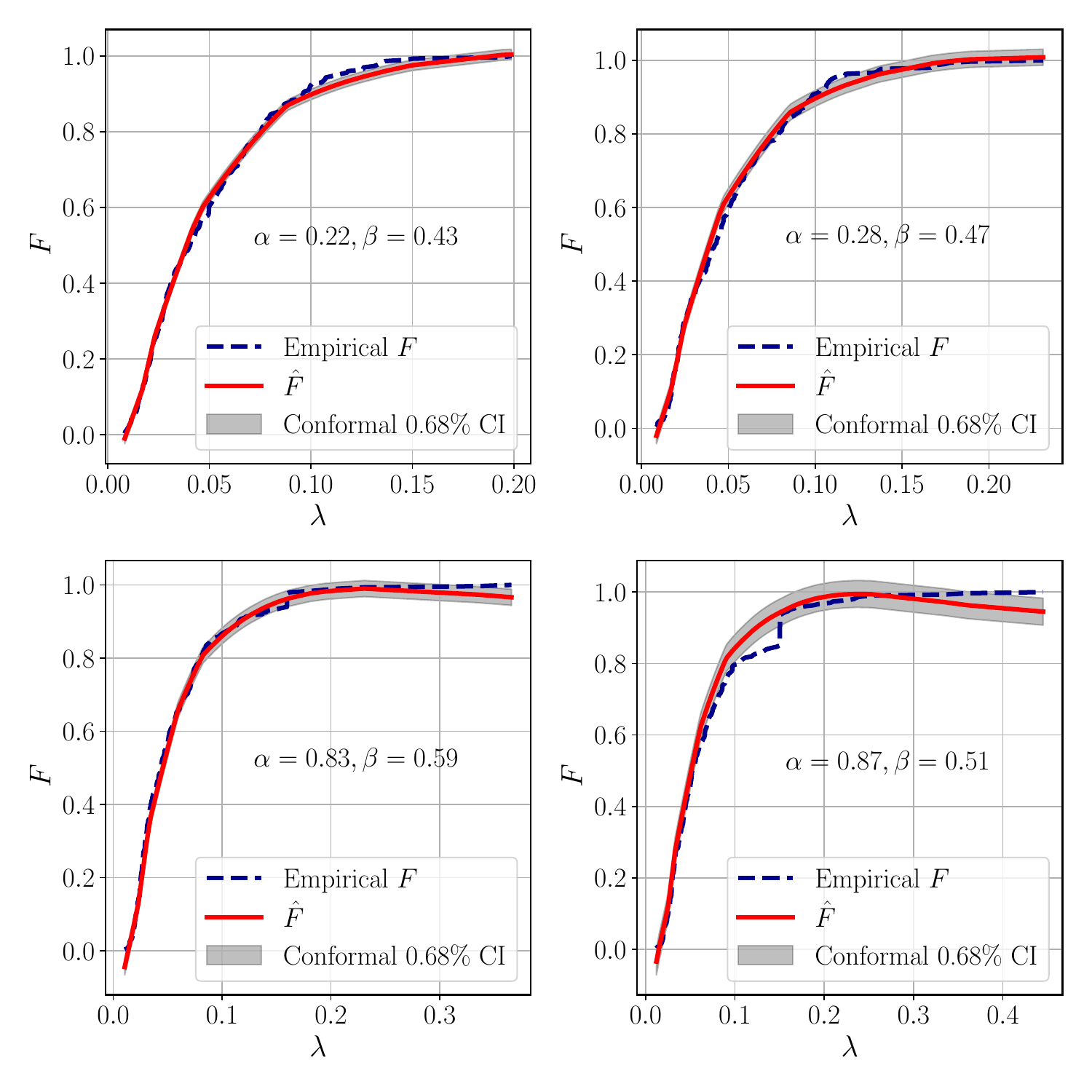}
    \caption{Cdfs for the optimized SIR model (MSE loss) obtained from modeling the empirical cdf at different $(\alpha,\beta)$ points with the associated $68\%$ conformal confidence band.}
    \label{fig:SIR_empiricalCDF_MSE}
  \end{minipage}
\end{figure*}

\begin{figure*}[h]
  \begin{minipage}[b]{\textwidth}
    \centering
\includegraphics[width=\textwidth,height=100mm]{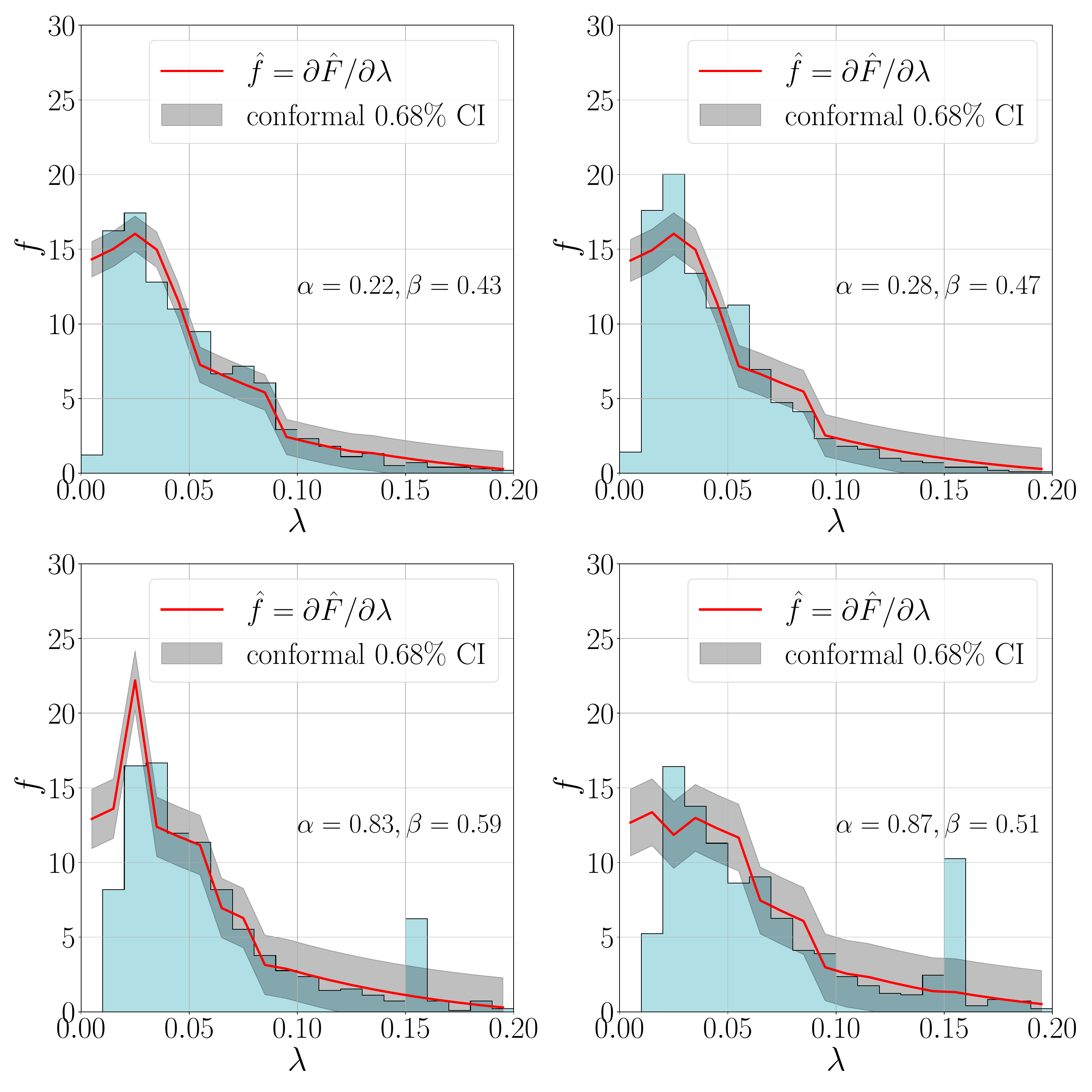}
    \caption{Pdfs for the optimized SIR model (MSE loss) obtained from modeling the empirical cdf at different $(\alpha,\beta)$ points with the associated $68\%$ conformal confidence band. The spikes around $\lambda \approx 0.15$ arise from simulated epidemics that die off within a day or so of the start of the simulated epidemics. In a considerably larger sample of simulated epidemics the spikes would not be visible.}
    \label{fig:SIR_PDF_MSE}
  \end{minipage}
\end{figure*}

\begin{figure*}[h]
  \centering
  \begin{minipage}[b]{\textwidth}
    \centering
    \includegraphics[width=\textwidth,height=100mm]{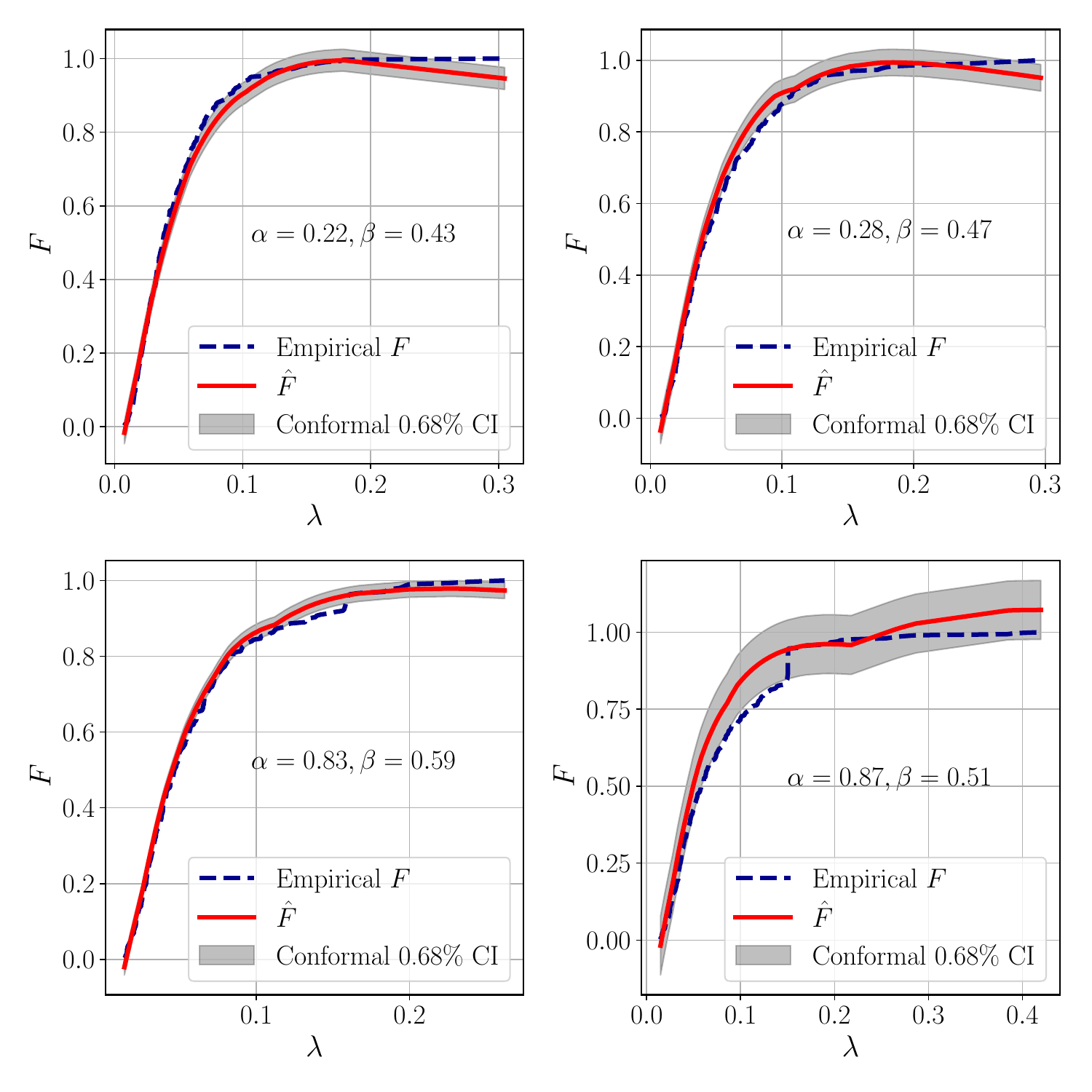}
    \caption{Cdfs for the optimized SIR model (Huber loss) obtained from modeling the empirical cdf at different $(\alpha,\beta)$ points with the associated $68\%$ conformal confidence band.}
    \label{fig:SIR_empiricalCDF_Huber}
  \end{minipage}
\end{figure*}

\begin{figure*}[h]
  \begin{minipage}[b]{\textwidth}
    \centering
\includegraphics[width=\textwidth,height=100mm]{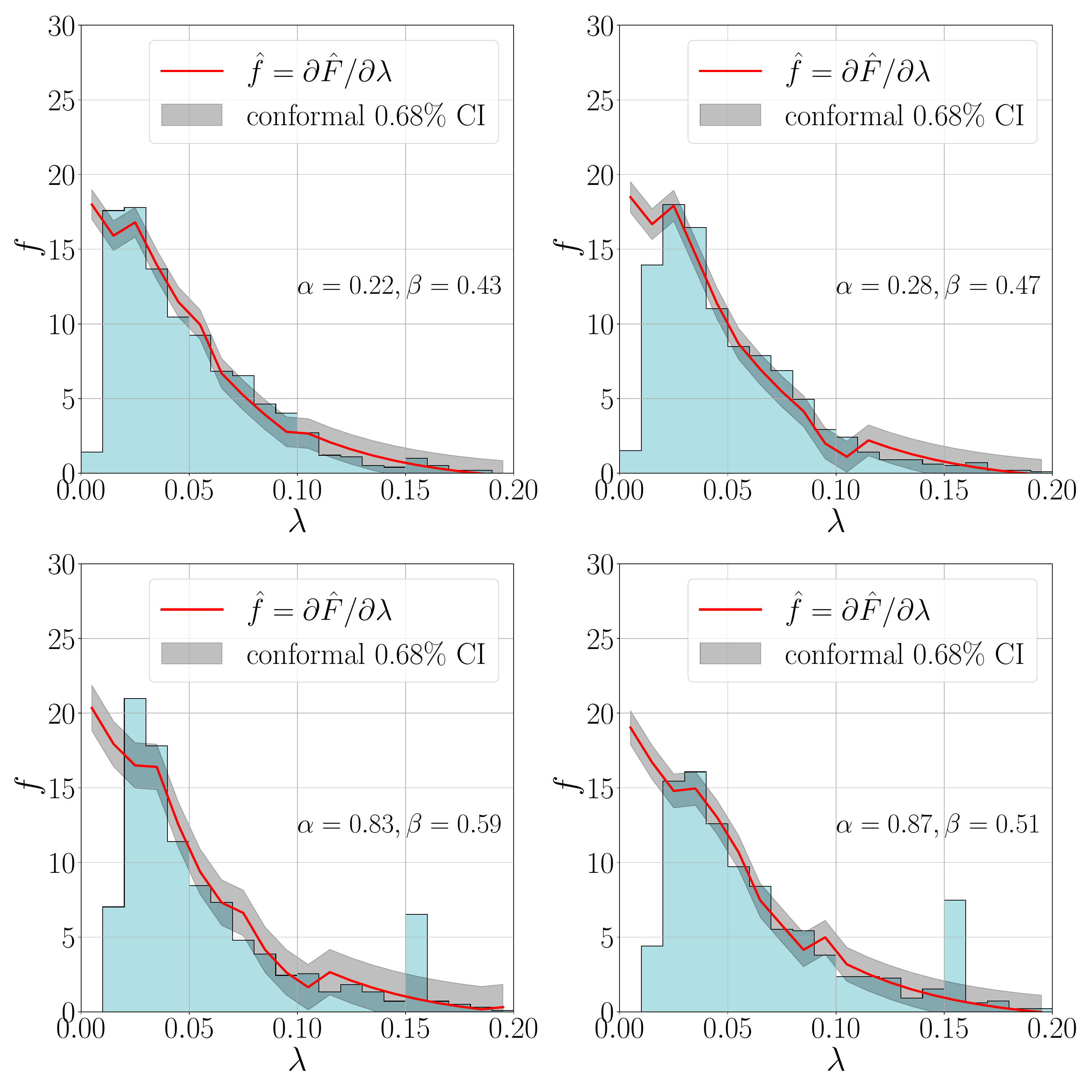}
    \caption{Pdfs for the optimized SIR model (Huber loss) obtained from modeling the empirical cdf at different $(\alpha,\beta)$ points with the associated $68\%$ conformal confidence band. The spikes around $\lambda \approx 0.15$ arise from simulated epidemics that die off within a day or so of the start of the simulated epidemics. In a considerably larger sample of simulated epidemics the spikes would not be visible.}
    \label{fig:SIR_PDF_Huber}
  \end{minipage}
\end{figure*}

Following the optimization process for our model hyperparameters, we initialized an MLP with the optimal hyperparameters identified for the MSE loss function. This network was then subjected to an extended training regimen of $2 \times 10^5$ iterations, equivalent to 39,960 epochs. Throughout training, we saved the model that achieved the lowest validation loss. 
Figures \ref{fig:SIR_empiricalCDF_MSE} and \ref{fig:SIR_PDF_MSE} depict the resulting estimated cdfs and pdfs with the 68\% conformal confidence intervals at different values of $\lambda$ and for different $(\alpha, \beta)$ points. To estimate the true pdf, which is necessary for the conformal inference algorithm, we performed a coarse-graining of $\lambda$. This entire procedure was repeated using a model trained with the Huber loss, using the optimal hyperparameters determined for the Huber loss. Figure \ref{fig:loss_curves} presents the training and validation loss curves over the course of training for both models, indicating that the Huber loss model achieves lower training and validation losses compared to the MSE loss model. Figures \ref{fig:SIR_empiricalCDF_Huber} and \ref{fig:SIR_PDF_Huber} display the resulting estimated cdfs and pdfs with the 68\% conformal confidence intervals at different values of $\lambda$ and for different $(\alpha, \beta)$ points. Pseudocode for the data generation, training, hyperparameter optimization, training and inference is shown in Algorithm \ref{alg:SIR_algorithm}. 

\subsection{Uncertainty Quantification using Bayesian Neural Networks}
Suppose we have a training dataset $\mathcal{D}=\{ \mathbf{x}_i, y_i \}_{i=1}^n$ (where $\mathbf{x}$ and $y$ are assumed to be drawn from a joint distribution $p(\mathbf{x},y)$). Our goal is to predict the target variable $y$ given a new input value $\mathbf{x}$. 
From a Bayesian perspective, the goal is to approximate the \textit{posterior predictive distribution} $p(y \mid \mathbf{x}, \mathcal{D})$ given new input $\mathbf{x}$ and training data $\mathcal{D}$,
\begin{equation}
\label{posterior_predictive}
    p(y \mid \mathbf{x}, \mathcal{D})=\int p(y \mid \mathbf{x}, \mathbf{w}) \, p(\mathbf{w} \mid \mathcal{D}) \, \mathrm{d} \mathbf{w},
\end{equation}
which entails a
marginalization over the network parameters. 
By Bayes' theorem, the posterior density over the neural network parameter space is given by

\begin{equation}
\label{posterior_weights}
    p(\mathbf{w} \mid \mathcal{D} ) = \frac{p(\mathcal{D} \mid \mathbf{w}) p(\mathbf{w})}{p(\mathcal{D})}, 
\end{equation}
which requires 
 the \textit{model evidence},
\begin{equation}
\label{eq:posterior_weights}
    p(\mathcal{D}) = \int p(\mathcal{D} \mid \mathbf{w} ) p(\mathbf{w}) d \mathbf{w}.
\end{equation}
Unlike the best-fit approach to neural networks, a Bayesian neural network (BNN) is the posterior density $p(\mathbf{w} \mid \mathcal{D} )$ over the network parameter space. 

The severe bottleneck with BNNs is computing the high-dimensional integrals Eq. \ref{posterior_predictive} and Eq. \ref{eq:posterior_weights}. In practice, these integrals must be approximated using various sampling schemes including Hamiltonian Monte Carlo \cite{Radford_Neal} or  variational inference \cite{Jordan_1999}. Furthermore, a prior density $p(\mathbf{w})$ over the neural network parameter space must be specified. Given a
collection of $K$ neural networks with parameters sampled from the posterior density, the predictive distribution can be approximated as follows
\begin{align}
    p(y \mid \mathbf{x}, \mathcal{D}) & = \frac{1}{K} \sum_{k=1}^K p(y | \mathbf{x}, \mathbf{w}_k),
\end{align}
where, typically, one models the density $p(y | \mathbf{x}, \mathbf{w}_k)$ as 
\begin{align}
    p(y | \mathbf{x}, \mathbf{w}_k) \approx \mathcal{N}(y; f(\mathbf{x}, \mathbf{w}), \sigma),
\end{align}
with $f(\mathbf{x}, \mathbf{w})$ the neural network.

\begin{figure*}[h]
  \centering
  \begin{minipage}[b]{\textwidth}
    \centering
    \includegraphics[width=\textwidth]{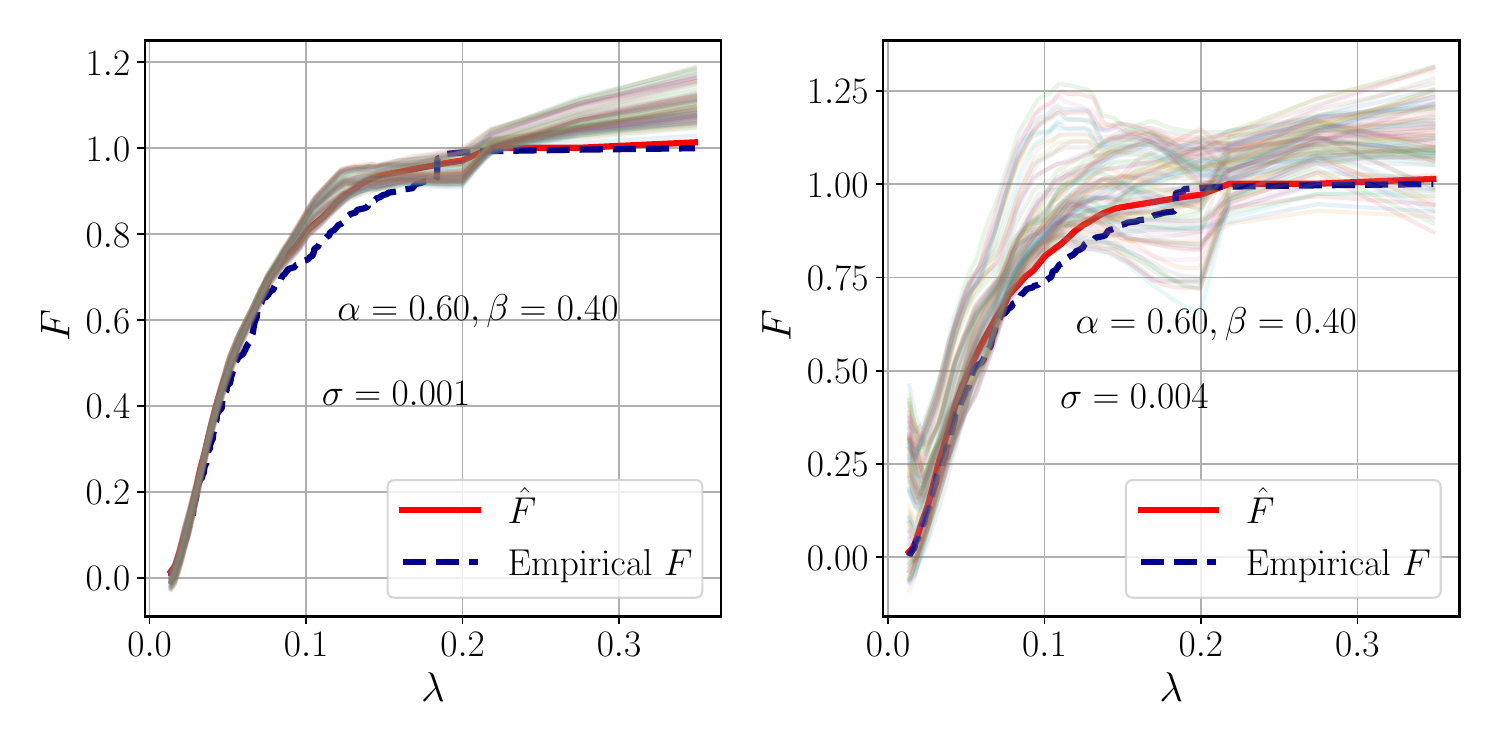}
    \caption{Fluctuating NN parameters by sampling from a prior modeled as a multivariate diagonal normal distribution with a single standard deviation $\sigma$. The spread of results is seen to be particularly sensitive to the choice of $\sigma$. We speculate that this could be due to the lack of adaptation of $\sigma$ to the different directions in the neural network parameter space.}
    \label{fig:fluctuate_Gaussian}
  \end{minipage}
\end{figure*}

In practice, given the computational burden of sampling from the true posterior density, we adopt a simpler approach.
Let $\mathbf{w^*} \in \mathbb{R}^k$ be the best-fit neural network parameters obtained by our training protocol using the Huber loss. We fluctuate the neural network parameters by sampling $\mathbf{w}_1,...,\mathbf{w}_n \sim \mathcal{N}(\mathbf{w^*}, \sigma^2 \mathbf{I}_k)$ for different choices of $\sigma$. With $n=100$, the resulting ensemble of models evaluated at the sampled parameters $\hat{f}(\mathbf{w}_1), ..., \hat{f}(\mathbf{w}_n)$ are shown in Fig. \ref{fig:fluctuate_Gaussian}.

\subsection{Bootstrap Neural Networks}
\label{sec:bootstrap_NN}
We follow the bootstrap idea (see Sec. \ref{sec:bootstrap})  to draw datasets with replacement from the training data, such that each bootstrap sample has the same size as the original training set.  This is done $K=200$ times, resulting in $K=200$ bootstrap training datasets. Due to practical considerations, we adopted a simplified neural network architecture consisting of 5 hidden layers, each containing 10 neurons, followed by a single output neuron utilizing a sigmoid activation function. The SiLU activation function was employed at each hidden node to enhance nonlinearity and learning capabilities. The network was trained using the NAdam optimizer with a fixed learning rate of $3 \times 10^{-4}$ in mini batches of size 60. This model architecture was retrained for $10^5$ iterations, with each model fitted to a different bootstrap sample.

\begin{figure*}[h]
    \centering
\includegraphics[width=0.7\textwidth,height=70mm]{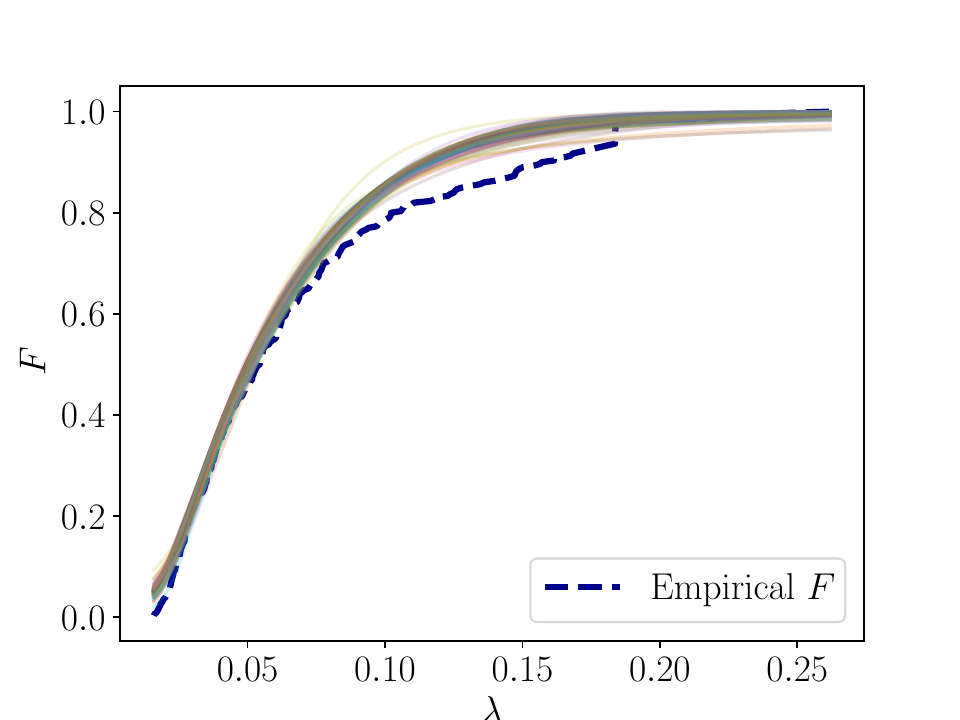}
    \caption{Responses of the 200 neural networks each trained on a different bootstrap training sample. See Sec. \ref{sec:bootstrap_NN} for details.}
  \label{fig:bootstrap}
\end{figure*}

The spread of the resulting models over the $K=200$ neural networks is displayed in Fig. \ref{fig:bootstrap}. The envelope of the spread is observed to be much smaller than for the previous methods. This could be due to insufficient training of the models. Other possible reasons for this are outlined below.

Although the bootstrap has been hailed as a satisfactory method for uncertainty quantification due to its theoretical guarantees \cite{RAYNER198843} it has limitations. For example, it is known to fail in situations where the data have a non-trivial dependency structure or when the functionals of interest are not smooth \cite{Bickel_Freedman}.
 Another question that was recently explored is the efficacy of bootstrap in the high-dimensional regime $n / d < 1$ relevant to machine learning,  where 
$d$ is the dimension of the machine learning model parameter space and $n$ the training sample size. In
\cite{karoui2016trust} it is shown that even for the simple case of inference in the linear model with unregularized least squares, bootstrap techniques perform poorly in only moderately high dimensions even with $n>d$.
It was also shown in \cite{clarté2024analysis} that resampling methods such as the bootstrap yield reliable error estimates only in the very low-dimensional regime $n \gg d$, and is fraught with problems in high-dimensional regimes. Therefore, perhaps one shouldn't be surprised by the results described above.

\section{Discussion and Conclusions}
\label{sec:discussion_conclusion}
Automatic differentiation is a fundamental tool in training neural networks, efficiently computing gradients with respect to model parameters and thus facilitating optimization algorithms like gradient descent. In this work, we exploit this capability to compute derivatives with respect to input variables just as is done, for example, in physics-informed neural networks (see for example \cite{PINN} and references therein). This allows one to investigate whether it is possible to accurately model a pdf by starting from a neural network model of the associated cdf. Cumulative distribution functions naturally arise in frequentist simulation-based inference.
One motivation for modeling the cdf $F(\lambda\mid\boldsymbol{\theta})$ as opposed to an \emph{ab initio} modeling of $f(\lambda \mid \boldsymbol{\theta})$ is that in general a better precision is reached approximating integrals of functions rather than the functions themselves. Furthermore, since the step of taking the derivative with autograd is exact, the accuracy of the approximated cdf directly translates to the accuracy of the desired pdf. 

It was found that a recently introduced method (\texttt{ALFFI}) that approximates the cdf as the mean of a certain discrete random variable appears not to be accurate enough in modeling the cdf for the purpose of deriving its associated pdf. We caution however that this may be the inevitable consequence of trying to model the cdf of an intrinsically discrete distribution with a smooth approximation. Our choice to investigate this distribution is intentional: it is a difficult test case for the proposed method.

We employ relatively simple yet effective NN models such that the amount of fine-tuning that is involved in going to more complex data settings is minimal. Several NN uncertainty quantification techniques were reviewed and implemented and we studied possible corrections of the cdf. Of the uncertainty quantification methods considered only one is calibrated by construction, namely, conformal inference.  We recommend that it be used as a simple benchmark with respect to which other uncertainty quantification methods can be compared and calibrated. 

As can be seen in Eq. \ref{validity} conformal inference results in confidence sets that guarantee \textit{marginal coverage}, $\mathbb{P}\left\{y_{n+1} \in \mathcal{C}_\alpha\left(x_{n+1}\right)\right\} \geq 1-\alpha$, that is, the coverage marginalized (integrated) over the population of which a dataset is a sample. 
That property is not at all controversial as coverage is a property of the population from which the dataset is presumed to have been drawn. But
if each element $x$ of the dataset contains a subset that are parameters $\theta$ sampled from a prior $\pi_\theta$ then the marginalization is also with respect to this prior. When a prior is involved 
this form of coverage is weaker than the more sought after \textit{conditional coverage} guarantee $\mathbb{P}\left\{y_{n+1} \in \mathcal{C}_\alpha\left(x_{n+1}\right) \mid \theta\right\} \geq 1-\alpha$. It was shown in \cite{Lei_Wasserman_2014} that such distribution-free conditional coverage is impossible to achieve with a finite sample, which has inspired studies that try to bridge the gap between marginal and conditional coverage \cite{gibbs2023conformal}. In our case, $x=\{ \theta,\lambda \}$ and we generate calibration sets at each $\theta$ point, thereby making our conformal confidence sets adaptive, that is, change with $\theta$. However, they are still not adaptive with respect to $\lambda$; the conformal confidence intervals are the same width at each $\lambda$ value. But it should be noted that this is not necessarily unusual. Consider, for example, the 68.3\% confidence interval $[x - \sigma, x + \sigma]$ of a Gaussian pdf. The widths of these intervals are twice the standard deviation $\sigma$ and, therefore, independent of $x$.

Another issue is that our conformal prediction sets are adaptive only because we have calibration data at the desired parameter points. It would be helpful to have a smooth interpolation of the width of the conformal confidence interval so that an interval can be computed at points where no calibration data exist.


\bmhead{Data Availability}
All data and code to reproduce all of our results is available at \url{https://github.com/hbprosper/cdf2pdf}.


\newpage
\begin{appendices}
\section{ALFFI algorithm for modeling CDF for ON/OFF Example}
\label{ALFFI_alg_OnOff}
\texttt{ALFFI} algorithm for modeling CDF in the ON/OFF problem, inspired by \cite{ALFFI}.

\begin{algorithm}[H]
\caption{Estimate the CDF $\mathbb{C}(\lambda \mid \mu, \nu)$, given the observed data $\{ N,M\}$ and the observed test statistic $\lambda$.}\label{p_val}
\begin{algorithmic}[1]
\Ensure estimated CDF $\widehat{\mathbb{C}}(\lambda \mid \mu, \nu)$ for all $\theta=\theta_0 \in \Theta$
\State set $\mathcal{T'} \gets \emptyset$
\For{$i$ in $\{1,...,B' \}$}
    \State Draw parameter $\mu_i \sim \pi_\mu = \text{Unif}(0,20)$
    \State Draw parameter $\nu_i \sim \pi_\nu = \text{Unif}(0,20)$
    \State Draw $n_i \sim \text{Poiss}(\mu_i+\nu_i)$
    \State Draw $m_i \sim \text{Poiss}(\nu_i)$
    \State Draw parameter $\mu_i' \sim \pi_{\mu'} = \text{Unif}(0,20)$
    \State Draw parameter $\nu_i' \sim \pi_{\nu'} = \text{Unif}(0,20)$

    \State Draw $N_i \sim \text{Poiss}(\mu_i'+\nu_i')$
    \State Draw $M_i \sim \text{Poiss}(\nu_i')$
    
    \State Compute test statistic under the null $\lambda_i \gets \lambda(n_i, m_i \mid \mu_i, \nu_i)$
    \State Compute indicator $Z_i \gets \mathbbm{1} ( \lambda_i \le \lambda(N_i, M_i \mid \mu_i, \nu_i) )$
    \State $\mathcal{T'} \gets \mathcal{T'} \cup \{ (\mu_i, \nu_i, \lambda_i, Z_i) \}$
\EndFor
\State Use $\mathcal{T'}$ to learn the parameterized function $\widehat{\mathbb{C}}(\lambda \mid \mu,\nu) := \mathbb{E}[Z \mid \mu, \nu ]$ via regression of $Z$ on $\{ \mu, \nu, \lambda \}$ using mean square error as the loss function
\State \Return $\widehat{\mathbb{C}}(\lambda \mid \mu, \nu)$, which estimates $\mathbb{P} (\lambda_i \le\lambda(N,M \mid \mu,\nu) )$
\end{algorithmic}
\end{algorithm}

\section{Residuals}

\begin{figure*}[h]
  \begin{minipage}[b]{\textwidth}
    \centering
    \includegraphics[width=\textwidth,height=100mm]{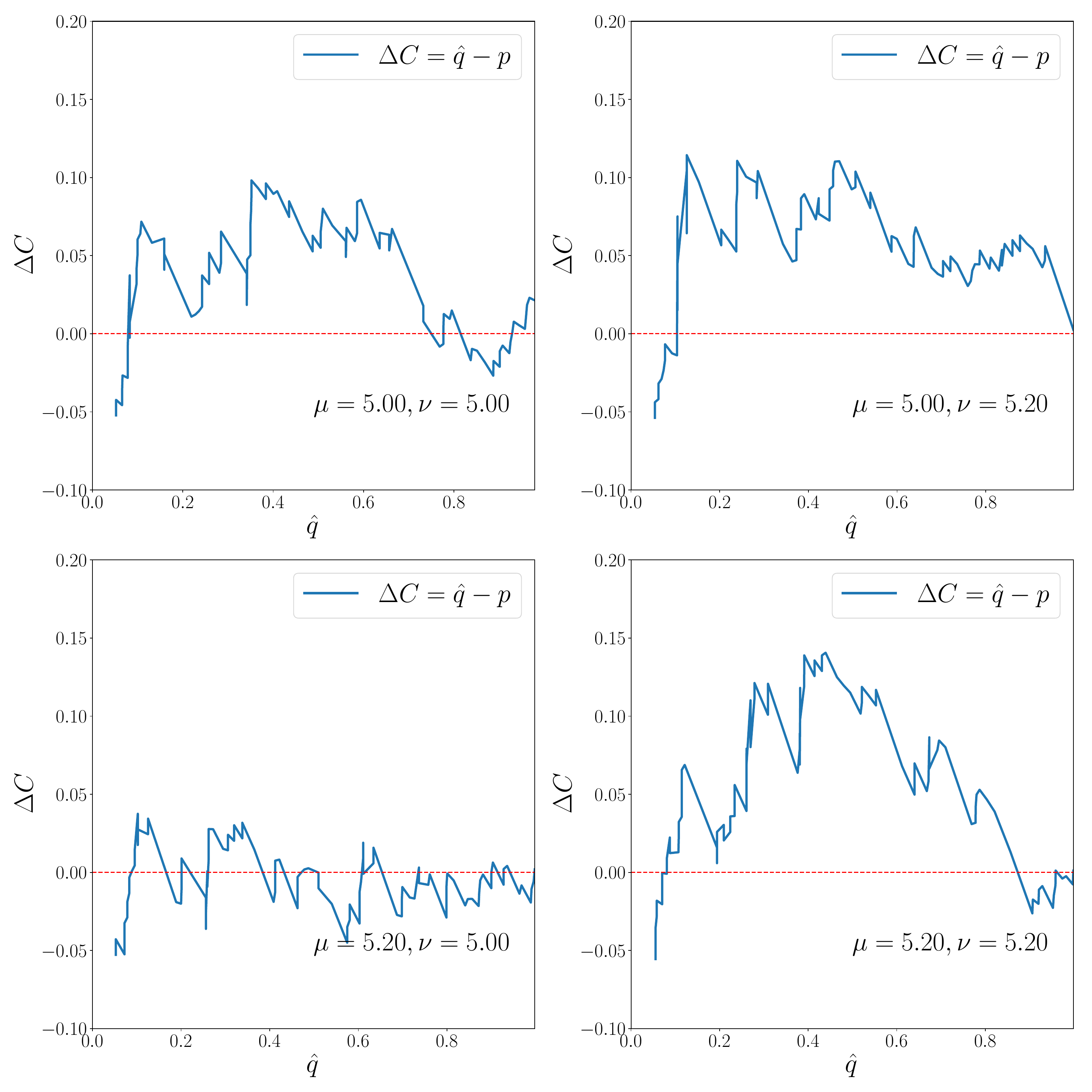}
    \caption{``quantile residuals" $\Delta C = \hat{q}-p$ as a function of $\hat{q}$ at different $(\mu,\nu)$ points.}
    \label{fig:deltaC}
  \end{minipage}
\end{figure*}

\begin{figure*}[h]
\label{chi2_conv}
  \centering
  \begin{minipage}[b]{\textwidth}
    \centering
    \includegraphics[width=\textwidth,height=100mm]{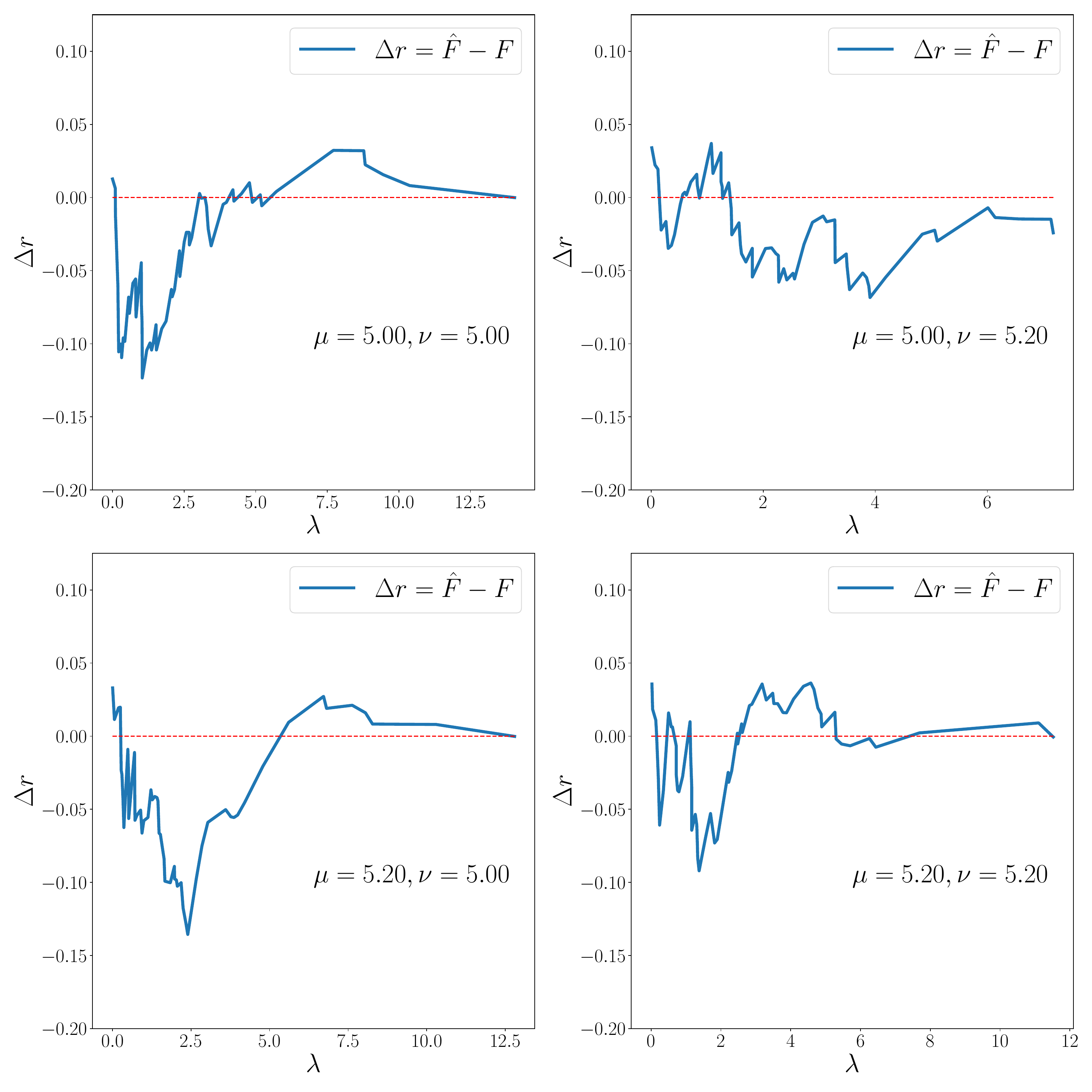}
    \caption{Cdf residuals $\Delta r = \hat{F}-F$ a a function of $\lambda$ at different $(\mu,\nu)$ points.}
    \label{fig:deltar}
  \end{minipage}
\end{figure*}

\end{appendices}

\newpage

\section*{Declarations}
\bmhead{Conflict of Interest}
All authors declare that they have no conflict of
interest.

\newpage
\bibliography{references}

\end{document}